%% file: main.tex
\documentclass[10pt,journal,compsoc]{IEEEtran}
\ifCLASSOPTIONcompsoc
  \usepackage[nocompress]{cite}
\else
  \usepackage{cite}
\fi
\ifCLASSINFOpdf
\else
\fi

\usepackage{epsfig}
\usepackage{graphicx}
\usepackage{amsmath}
\usepackage{amssymb}
\usepackage{enumitem}

\usepackage[pagebackref=true,breaklinks=true,letterpaper=true,colorlinks,bookmarks=false,urlcolor=red,citecolor=cyan]{hyperref}
\usepackage{gensymb,graphics,subfigure,inputenc}
\usepackage[linesnumbered,algo2e,boxed]{algorithm2e}
\graphicspath{{Figure/}}
\usepackage[figuresright]{rotating}
\usepackage{amsmath,amssymb,textcomp,subfigure,multirow,upgreek}
\usepackage{booktabs}
\usepackage{color,mdwlist}

\usepackage{ragged2e}

\usepackage{review}

\usepackage{graphicx,fontawesome}
\graphicspath{{Figures/}}

\usepackage{caption}
\usepackage{wrapfig}

\usepackage{enumitem}
\setitemize{noitemsep,topsep=0pt,parsep=0pt,partopsep=0pt}

\begin{document}
%
\title{Local and Global GANs with Semantic-Aware Upsampling for Image Generation}
\author{Hao~Tang,
    Ling Shao,
 	Philip H.S. Torr,
	Nicu~Sebe
	\IEEEcompsocitemizethanks{
	    \IEEEcompsocthanksitem Hao Tang is with the Department of Information Technology and Electrical Engineering, ETH Zurich,  Zurich 8092, Switzerland. E-mail: hao.tang@vision.ee.ethz.ch \protect
	    \IEEEcompsocthanksitem Ling Shao is with the Inception Institute of Artificial Intelligence, UAE. \protect
	    \IEEEcompsocthanksitem  Philip H.S. Torr is with the Department of Engineering Science, University of Oxford, UK. \protect
	    \IEEEcompsocthanksitem Nicu Sebe is with the Department of Information Engineering and Computer Science (DISI), University of Trento, Italy. \protect
        }
	\thanks{Corresponding author: Hao Tang.}
}

%
%

\markboth{IEEE Transactions on Pattern Analysis and Machine Intelligence}%
{Shell \MakeLowercase{\textit{et al.}}: Bare Demo of IEEEtran.cls for Computer Society Journals}
%



\IEEEtitleabstractindextext{%
\input{0Abstract}

}

\maketitle

\IEEEdisplaynontitleabstractindextext

%
\IEEEpeerreviewmaketitle


%
%
%
%

\input{1Introduction}
\input{2RelatedWork}
\input{3Formulation}
\input{4Implementation}
\input{5Conclusion}
\input{6appendices}
\footnotesize
\bibliographystyle{IEEEtran}
\bibliography{egbib}

\begin{IEEEbiography}[{\includegraphics[width=1in,height=1.25in,clip,keepaspectratio]{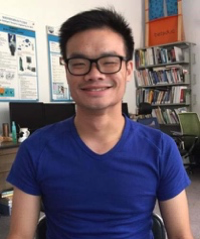}}]{Hao Tang} is currently a Postdoctoral with Computer Vision Lab, ETH Zurich, Switzerland.
He received the master’s degree from the School of Electronics and Computer Engineering, Peking University, China and the Ph.D. degree from Multimedia and Human Understanding Group, University of Trento, Italy.
He was a visiting scholar in the Department of Engineering Science at the University of Oxford. His research interests are deep learning, machine learning, and their applications to computer vision.
\end{IEEEbiography}

\begin{IEEEbiography}[{\includegraphics[width=1in,height=1.25in,clip,keepaspectratio]{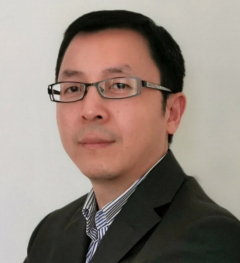}}]{Ling Shao} is the CEO of the Inception Institute of Artificial Intelligence, and the Executive Vice President and Provost of the Mohamed bin Zayed University of Artificial Intelligence, Abu Dhabi, UAE. 
He received the B.Eng. degree in Electronic and Information Engineering from the University of Science and Technology of China (USTC), the M.Sc degree in Medical Image Analysis and the PhD degree in Computer Vision at the Robotics Research Group from the University of Oxford. 
	
\end{IEEEbiography}

\begin{IEEEbiography}[{\includegraphics[width=1in,height=1.25in,clip,keepaspectratio]{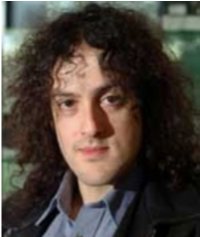}}]{Philip H. S. Torr} received the PhD degree from Oxford University. After working for another three years at Oxford, he worked for six years for
	Microsoft Research, first in Redmond, then in
	Cambridge, founding the vision side of the Machine Learning and Perception Group. He is now
	a professor at Oxford University. He has won
	awards from top vision conferences, including
	ICCV, CVPR, ECCV, NIPS and BMVC. He is a
	senior member of the IEEE and a Royal Society
	Wolfson Research Merit Award holder.
\end{IEEEbiography}

\begin{IEEEbiography}[{\includegraphics[width=1in,height=1.25in,clip,keepaspectratio]{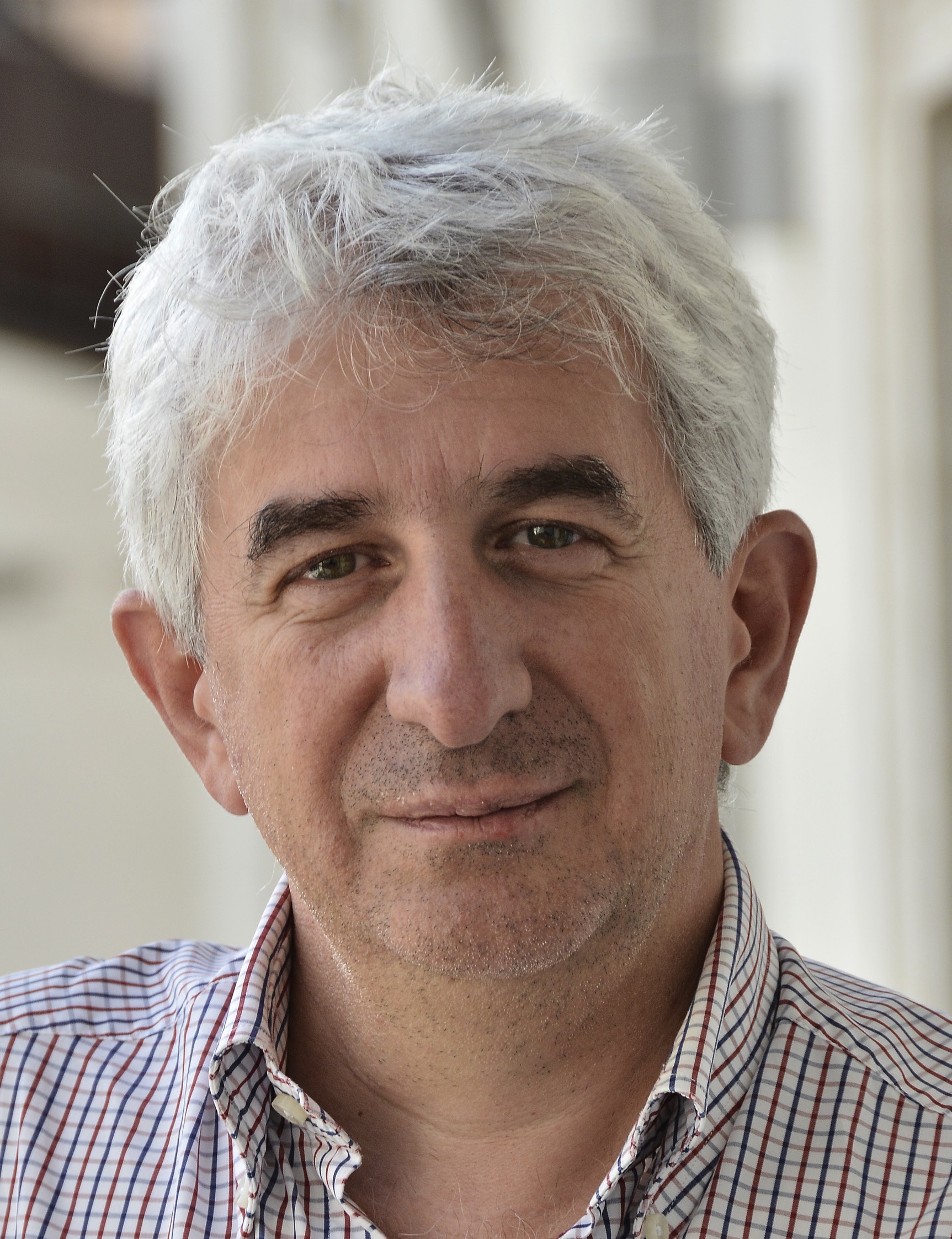}}]{Nicu Sebe} is Professor with the University of Trento, Italy, leading the research in the areas of multimedia information retrieval and human behavior understanding. He was the General Co-Chair of ACM Multimedia 2013, and the Program Chair of ACM Multimedia 2007 and 2011, ECCV 2016, ICCV 2017 and ICPR 2020. He is a fellow of the International Association for Pattern Recognition.
\end{IEEEbiography}




\end{document}

%% file: 0Abstract.tex
\justify
\begin{abstract}
In this paper, we address the task of semantic-guided image generation. One challenge common to most existing image-level generation methods is the difficulty in generating small objects and detailed local textures. To address this, in this work we consider generating images using local context. As such, we design a local class-specific generative network using semantic maps as guidance, which separately constructs and learns subgenerators for different classes, enabling it to capture finer details. To learn more discriminative class-specific feature representations for the local generation, we also propose a novel classification module. To combine the advantages of both global image-level and local class-specific generation, a joint generation network is designed with an attention fusion module and a dual-discriminator structure embedded.
Lastly, we propose a novel semantic-aware upsampling method, which has a larger receptive field and can take far-away pixels that are semantically related for feature upsampling, enabling it to better preserve semantic consistency for instances with the same semantic labels. Extensive experiments on two image generation tasks show the superior performance of the proposed method. State-of-the-art results are established by large margins on both tasks and on nine challenging public benchmarks. The source code and trained models are available at
\url{https://github.com/Ha0Tang/LGGAN}.
\end{abstract}

%% file: 1Introduction.tex
\section{Introduction}
\begin{figure}[!t] \small
	\centering
	\includegraphics[width=1\linewidth]{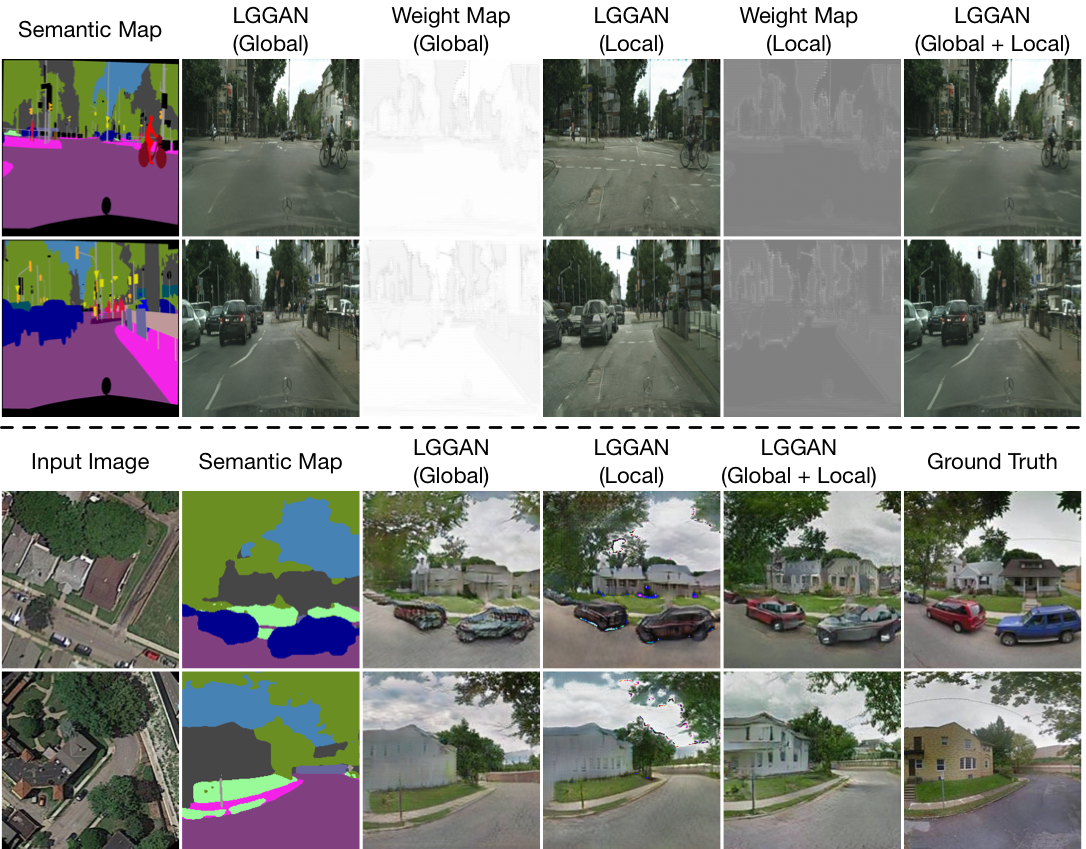}
	\caption{Examples of semantic image synthesis results on Cityscapes (top) and cross-view image translation results on Dayton (bottom) with different settings of our LGGAN.}
	\label{fig:first}
	\vspace{-0.4cm}
\end{figure}

\IEEEPARstart{S}{emantic}-guided image generation is a hot topic covering several mainstream research directions, including cross-view image translation \cite{isola2017image,zhai2017predicting,regmi2018cross,tang2019multi} and semantic image synthesis~\cite{wang2018high, chen2017photographic,qi2018semi, park2019semantic,zhu2020semantically,zhu2020sean}. 
The cross-view image translation task, proposed in \cite{regmi2018cross}, is essentially an ill-posed problem due to the large ambiguity in the generation if only a single RGB image is given as input. 
To alleviate this problem, recent works, such as SelectionGAN~\cite{tang2019multi}, try to generate the target image based on another image of the scene and several novel semantic maps, as shown in Fig.~\ref{fig:first} (bottom).
Adding a semantic map allows the model to learn the correspondences in the target view with appropriate object relations and transformations.
On the other side, the semantic image synthesis task aims to generate a photorealistic image from a semantic map (see Fig.~\ref{fig:first} (top)).
This has many real-world applications and has drawn much attention from the academic research community as well as industry \cite{wang2018high, chen2017photographic, qi2018semi, park2019semantic,liu2019learning,jiang2020tsit,tang2020dual}. 
With the use of semantic information, existing methods for both tasks have achieved promising performance in semantic-guided image generation.

\begin{figure*}[!t] \small
	\centering
	\includegraphics[width=1\linewidth]{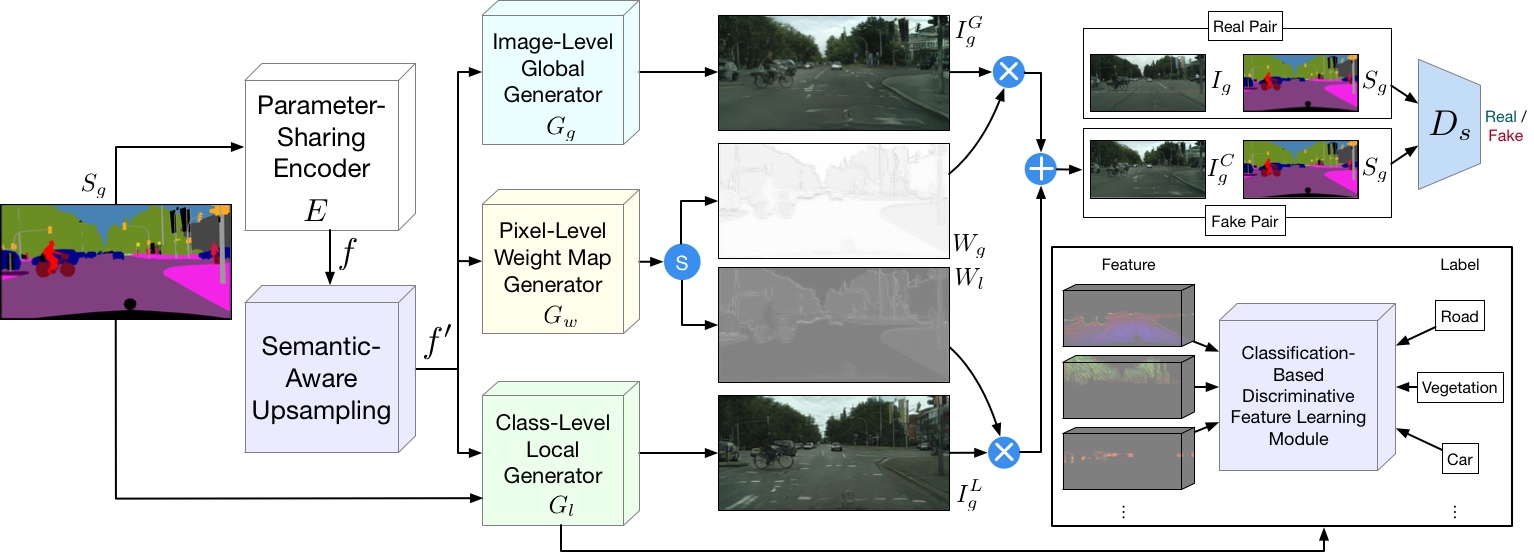}
	\caption{Overview of the proposed method, which contains a semantic-guided generator $G$ and discriminator $D_s$. $G$ consists of a parameter-sharing encoder $E$, an image-level global generator $G_g$, a class-level local generator $G_l$ and a weight map generator $G_w$.
	The global generator and local generator are automatically combined by two learned weight maps from the weight map generator to reconstruct the target image.
	$D_s$ tries to distinguish the generated images from two modality spaces, i.e., the image space and semantic space. 
	Moreover, to learn a more discriminative class-specific feature representation, a novel classification module is proposed.
	Lastly, to better preserve semantic information when upsampling feature maps, a novel semantic-aware upsampling method is introduced.
	All of these components are trained in an end-to-end fashion so that the local generation and the global generation can benefit from each other.
	The symbols $\oplus$, $\otimes$ and $\textcircled{s}$ denote element-wise addition, element-wise multiplication and channel-wise Softmax, respectively.}
	\label{fig:framework}
	\vspace{-0.4cm}
\end{figure*}

However, there is still room for improvement, especially when it comes to generating local structures and details, as well as small-scale objects. 
We believe there are several reasons for this. 
First, the existing methods for both tasks are typically based on global image-level generation.
In other words, they accept a semantic map containing several object classes and aim to generate the appearance of each one using the same network design or shared network parameters. In this case, all the classes are treated equally by the network. 
However, because different semantic classes have distinct properties, using specified network learning for each would intuitively facilitate the complex generation of multiple classes. 
Second, the number of training samples for different semantic classes is often imbalanced.
For instance, for the Dayton dataset~\cite{vo2016localizing}, cars and buses occupy less than 2\% of all pixels in the training data.
This naturally causes the model learning to be dominated by the classes with the largest number of training samples. 
Third, the sizes of objects in different semantic classes vary. 
As shown in the first row of Fig.~\ref{fig:first}, larger-scale object classes, such as roads and sky, usually occupy a bigger area of the image than smaller-scale classes, such as poles and traffic lights. Since convolutional networks usually share parameters at different convolutional positions, the larger-scale object classes would thus take advantage during the learning process, further increasing the difficulty in accurately generating the small-scale object classes.

To tackle these issues, a straightforward solution would be to model the generation of different image classes individually using local context. By so doing, each class could have its own generation network structure or parameters, thus greatly avoiding the learning of a biased generation space. To achieve this goal, in this paper, we design a novel class-specific generation network. It consists of several subgenerators for different classes with a shared encoded feature map. The input semantic map is utilized as the guidance to obtain feature maps corresponding spatially to each class, which are then used to produce a separate generation for different class regions.  

Due to the highly complementary properties of global and local generation, a local class-specific and global image-level generative adversarial network~(LGGAN) is proposed to combine the advantages of both. It contains three main network branches (see Fig.~\ref{fig:framework}). 
The first is the image-level global generator, which learns a global appearance distribution using the input. 
The second is the proposed class-specific local generator, which aims to generate different object classes separately, using semantic-guided class-specific feature filtering. Finally, the fusion weight map generation branch learns two pixel-level weight maps, which are used to fuse the local and global subnetworks in a weighted combination of their final generation results. The proposed LGGAN can be jointly trained in an end-to-end fashion, enabling the local and global generation to benefit from each other during optimization. 

\begin{figure*}[!t]\small
\centering
\subfigure[Input Semantic Map]{\label{fig:m1}\includegraphics[width=0.195\linewidth]{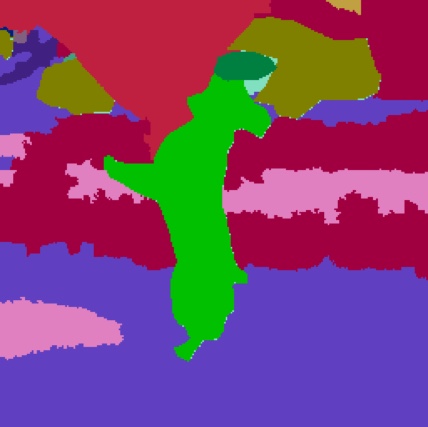}}
\subfigure[Nearest, bilinear, etc]{\label{fig:m2}\includegraphics[width=0.195\linewidth]{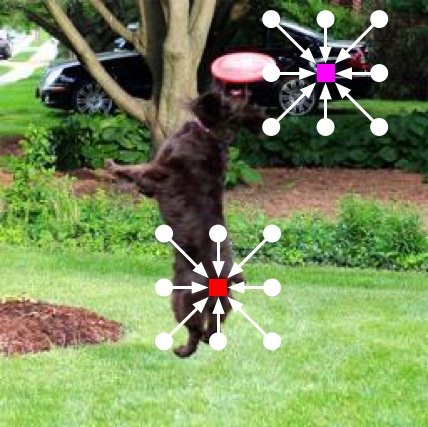}}
\subfigure[Deconvolution]{\label{fig:m3}\includegraphics[width=0.195\linewidth]{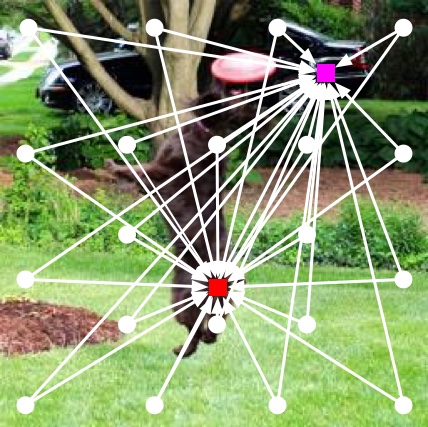}}
\subfigure[Spatial Attention]{\label{fig:m4}\includegraphics[width=0.195\linewidth]{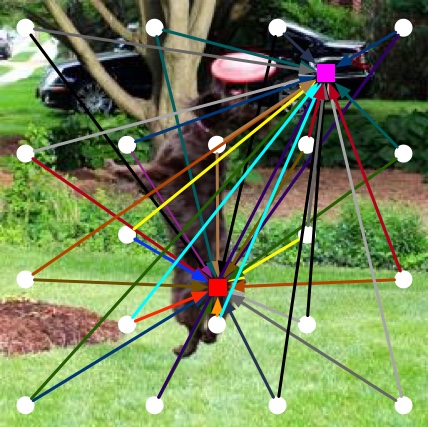}}
\subfigure[SAU (Ours)]{\label{fig:m5}\includegraphics[width=0.195\linewidth]{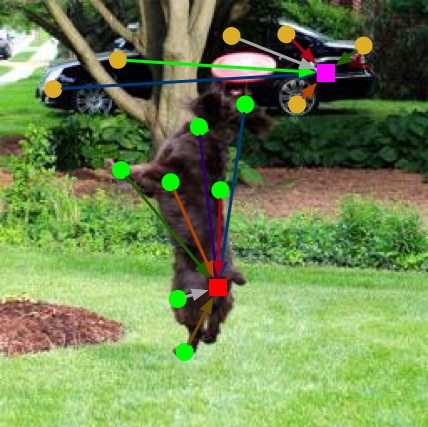}}
\caption{Comparison with different feature upsampling and enhancement methods on the semantic-guided image generation task.
Given two locations (indicated by red and magenta squares) in the output feature map, our goal is to generate these locations by selectively upsampling several points (indicated by circles) in the input feature map.}
\label{fig:motivation}
	\vspace{-0.4cm}
\end{figure*}

Moreover, existing methods, such as \cite{park2019semantic,tang2020dual}, typically adopt nearest-neighbor interpolation to upsample feature maps and then generate final results, which leads to many visual artifacts and blurriness in the generated images. 
Feature upsampling is a key operation in the semantic-guided image synthesis task.  
Traditional upsampling methods, such as nearest-neighbor, bilinear, and bicubic interpolation only consider subpixel neighborhoods (indicated by white circles in Fig.~\ref{fig:m2}), failing to capture the complete semantic information, e.g., the head and body of the dog, and the front part of the car.
Learnable upsampling methods, such as deconvolution \cite{noh2015learning} and pixel shuffle \cite{shi2016real}, are able to obtain the global information with a larger kernel size, but learn the same kernel (indicated by the white arrows in Fig.~\ref{fig:m3}) across the image, regardless of the semantic information. 
Other feature enhancement methods, such as spatial attention \cite{fu2019dual}, can learn different kernels (indicated by different colored arrows in Fig.~\ref{fig:m4}), but they still inevitably capture a lot of redundant information, i.e., `grass' and `soil'.
Moreover, they are prohibitively expensive since they must consider all pixels.

To address these limitations, we further propose a novel semantic-aware upsampling (SAU) for this challenging task, as shown in Fig.~\ref{fig:m5}.
Our SAU dynamically upsamples a small subset of relevant pixels based on the semantic information, i.e., the green and tangerine circles represent the pixels within the dog and the car, respectively.
In this way, SAU is more efficient than deconvolution, pixel shuffle, and spatial attention, and can capture more complete semantic information than traditional upsampling methods such as nearest-neighbor interpolation.

Overall, the contributions of this paper are as follows:
\begin{itemize}[leftmargin=*]
	\item We explore image generation from local context, which we believe is beneficial for generating richer details compared with the existing global image-level generation methods. A new local class-specific generative structure is designed for this purpose. It can effectively handle the generation of small objects and details, which are common difficulties encountered by the global-based generation.
	\item We propose a novel global and local generative adversarial network design able to take into account both the global and local contexts. To stabilize the optimization of the proposed joint network structure, a fusion weight map generator and a dual-discriminator are introduced. Moreover, to learn discriminative class-specific feature representations, a novel classification module is proposed.
	\item We introduce a novel semantic-aware upsampling (SAU) to dynamically upsample a small subset of relevant pixels based on the semantic information. SAU is more efficient than deconvolution, pixel shuffle, and spatial attention, and can capture more complete semantic information than traditional upsampling methods such as nearest-neighbor interpolation.
	\item Experiments for cross-view image translation on the Dayton~\cite{vo2016localizing}, CVUSA~\cite{workman2015wide}, and SVA~\cite{palazzi2017learning} datasets, and semantic image synthesis on the Cityscapes \cite{cordts2016cityscapes}, ADE20K \cite{zhou2017scene}, COCO-Stuff \cite{caesar2018coco}, DeepFashion \cite{liu2016deepfashion}, CelebAMask-HQ \cite{CelebAMask-HQ}, and Facades \cite{tylevcek2013spatial} datasets demonstrate the effectiveness of the proposed method, and show significantly better results compared with state-of-the-art methods. 
\end{itemize}

Part of the material presented here appeared in \cite{tang2020local}. The current paper extends \cite{tang2020local} in several ways.
(1) We present a more detailed analysis of related works, including recently published works dealing with semantic-guided image generation and feature upsampling. 
(2) We propose a general and highly effective feature upsampling method, i.e., SAU. The proposed SAU has three advantages: i) Global view. Unlike traditional upsampling methods (e.g., nearest-neighbor) that only exploit local neighborhoods, SAU can aggregate semantic information in a global view. ii) Semantically adaptive. 
Instead of using a fixed kernel for all locations (e.g., deconvolution), SAU enables semantic class-specific upsampling by generating adaptive kernels for different locations. iii) Efficient. Unlike spatial attention, which uses a fully connected strategy to connect all pixels, SAU only considers the most relevant pixels, introducing little computational overhead. 
Equipped with this new upsampling method, our LGGAN proposed in \cite{tang2020local} is upgraded to LGGAN++.
(3) We conduct extensive ablation studies to demonstrate the effectiveness of the proposed SAU against other feature upsampling and enhancement methods.  
(4) We extend the quantitative and qualitative experiments by comparing the proposed LGGAN and SAU with very recent works on two image synthesis tasks with diverse scenarios. We observe that the proposed methods achieve consistent and substantial gains on nine public datasets.

%% file: 2RelatedWork.tex
\section{Related Work}

\noindent\textbf{Generative Adversarial Networks (GANs)}
\cite{goodfellow2014generative} have been widely used for image generation \cite{karras2018style,shaham2019singan}.
A vanilla GAN has two important components, i.e., a generator and a discriminator. 
The goal of the generator is to synthesize photorealistic images from a noise vector, while the discriminator tries to distinguish between the real and the generated images. 
To create user-specific images, the conditional GAN (CGAN) \cite{mirza2014conditional} was proposed.
A CGAN combines a vanilla GAN and external information, such as class labels~\cite{odena2016conditional,choi2017stargan}, text descriptions \cite{han2017stackgan,li2019controllable}, object/face keypoints~\cite{reed2016learning,tang2019cycle}, human body/hand skeletons~\cite{tang2020xinggan,tang2018gesturegan,tang2021total,tang2020unified,tang2020bipartite}, semantic maps~\cite{wang2018high,tang2019multi}, or attention maps~\cite{mejjati2018unsupervised,tang2021attentiongan}.

\noindent\textbf{Global and Local Generation in GANs.}
Modeling global and local information in GANs to generate better results has been explored in various generative tasks \cite{huang2017beyond,iizuka2017globally,lin2019coco,li2018global,qi2018global,gu2019mask}. 
For instance, Huang et al.~\cite{huang2017beyond} proposed TPGAN for frontal view synthesis by simultaneously perceiving global structures and local details.
Gu et al.~\cite{gu2019mask} proposed MaskGAN for face editing by separately learning every face component, e.g., mouth and eye.
However, these methods have only been applied to face-related tasks, such as face rotation or face editing, where all the domains have large overlap and similarity.
In contrast, we propose a new local and global image generation framework for the more challenging scene image generation task, where the local context modeling is based on semantic-guided class-specific generation, which has not been explored by any existing works.

\begin{figure*} [!t] \small
	\centering
	\includegraphics[width=1\linewidth]{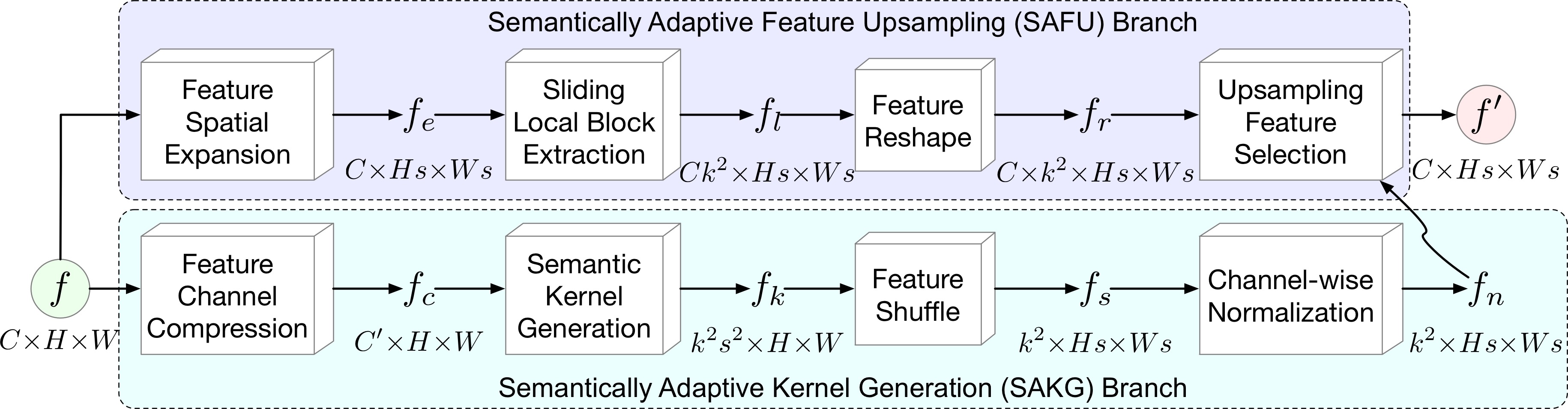}
	\caption{Overview of the proposed SAU, which consists of two branches, i.e., SAKG and SAFU. 
	The SAKG branch aims to generate semantically adaptive kernels according to the input layout.
	The SAFU branch aims to selectively upsample the feature $f{\in} C {\times} H {\times} W$ to the target one $f'{\in} C {\times} {Hs} {\times} {Ws}$ based on the kernels learned in SAKG, where $s$ is the expected upsample scale.}
	\label{fig:method}
	\vspace{-0.4cm}
\end{figure*}

\noindent\textbf{Semantic-Guided Image Generation.}
Scene generation tasks are a hot topic as each image can be parsed into distinctive semantic objects.
In this paper, we mainly focus on two image generation tasks, i.e., cross-view image translation \cite{zhai2017predicting,regmi2018cross,regmi2019cross,tang2019multi,ren2021cascaded} and semantic image synthesis \cite{wang2018high,chen2017photographic,qi2018semi, park2019semantic,tang2021layout}.
Most existing works on cross-view image translation aim to synthesize novel views of the same objects \cite{dosovitskiy2017learning,zhou2016view,tatarchenko2016multi}.
For instance, Dosovitskiy et al. \cite{dosovitskiy2017learning} used generative models to produce unseen views of cars, chairs, and tables.
Moreover, several works deal with image translation problems with drastically different views and generate a novel scene from another given scene ~\cite{zhai2017predicting,regmi2018cross,tang2019multi}.
For instance, Zhai et al. \cite{zhai2017predicting} tried to generate panoramic ground-level images from aerial images of the same location.
Tang~et al. proposed SelectionGAN \cite{tang2019multi} to solve the cross-view image translation task using semantic maps and CGAN models.
On the other side, the semantic image synthesis task aims to generate a photorealistic image from a semantic map~\cite{wang2018high, chen2017photographic, park2019semantic}.
For example, Park et al. proposed GauGAN~\cite{park2019semantic}, which achieves the best results on this task.

With the semantic maps as guidance, existing approaches for both tasks achieve promising performance.
However, the results produced by these global image-level generation methods are still often unsatisfactory, especially for detailed local textures.
In contrast, our proposed approach focuses on generating a more realistic global structure/layout and local texture details.
Both the local and global generation branches are jointly learned in an end-to-end fashion, enabling both to be improved through their mutually beneficial information.

\noindent \textbf{Feature Upsampling.} 
Traditional upsampling methods such as nearest-neighbor and bilinear interpolation use spatial distance and handcrafted kernels to capture the correlations between pixels. 
Recently, several deep learning methods, such as deconvolution \cite{noh2015learning} and pixel shuffle \cite{shi2016real}, have been proposed to upsample feature maps using learnable kernels.
For instance, pixel shuffle \cite{shi2016real} tries to reshape depth on the channel dimension into width and height on the spatial dimension.
However, these methods either exploit semantic information in a small neighborhood or use a fixed kernel.
Other works, including super-resolution \cite{jo2018deep,hu2019meta}, inpainting \cite{wang2019carafe}, and denoising \cite{mildenhall2018burst}, also explore the use of learnable kernels.

Existing image generation methods, such as \cite{park2019semantic,jiang2020tsit,tang2020dual}, typically adopt nearest-neighbor interpolation to upsample feature maps and then generate final results. 
However, this leads to unsatisfactory results, particularly in the generated content details and intra-object completions.
To address this limitation, we propose a novel semantic-aware upsampling for this task.
To the best of our knowledge, we are the first to investigate the influence of feature upsampling on this challenging task.

%% file: 3Formulation.tex
\section{The Proposed SAU and LGGAN}
We start by presenting the details of the proposed SAU and LGGAN. 
We first introduce the used backbone structure and then present the proposed semantic-aware upsampling, and finally introduce the design of the proposed local and global generation networks.

\subsection{Backbone Encoding Network Structure}
\noindent \textbf{Semantic-Guided Generation.}
In this paper, we focus on two main tasks, i.e., semantic image synthesis and cross-view image translation.
For the former, we follow GauGAN~\cite{park2019semantic} and use the semantic map $S_g$ as the input of the backbone encoder $E$, as shown in Fig.~\ref{fig:framework}.
For the latter, we follow SelectionGAN~\cite{tang2019multi} and concatenate the input image $I_a$ and a novel semantic map $S_g$ as the input of the backbone encoder $E$.
By so doing, the semantic maps act as priors to guide the model in learning the generation of another domain.

\noindent \textbf{Parameter-Sharing Encoder.}
As we have three different branches for three different generators, the encoder $E$ shares parameters with all the branches to create a compact backbone network. 
The gradients from each branch all contribute to the learning of the encoder. 
We believe that, in this way, the encoder can learn both local and global information and the correspondence between them.
Thus, the encoded deep representation of the input $S_g$ can be represented as $f{=}E(S_g)$.
This feature $f$ is then fed into the proposed SAU to obtain an upsampled feature map $f'{=}{\rm SAU}(f)$, as shown in Fig.~\ref{fig:framework} and \ref{fig:method}. 

\subsection{Semantic-Aware Upsampling}

An illustration of the proposed SAU is shown in Fig.~\ref{fig:method}.
It consists of two main branches, i.e., the semantically adaptive kernel generation (SAKG) branch, which predicts upsampled kernels according to the semantic information, and the semantically adaptive feature upsampling (SAFU) branch, which selectively performs the feature upsampling based on the kernels learned in SAKG. All components are trained in an end-to-end fashion.

Specifically, given a feature map $f{\in} \mathbb{R}^{C{\times} H {\times} W}$ and an upsample scale $s$, SAU aims to produce a new feature map $f'{\in} \mathbb{R}^{C{\times} Hs {\times} Ws}$. 
For any target location $l'{=} (i',j')$ in the output $f'$, there is a corresponding source location $l{=} (i,j)$ at the input $f$, where $i{=}\lfloor i' {/} s\rfloor$, $j{=}\lfloor j' {/} s\rfloor$.
We denote $Z(l, k)$ as the $k{\times}k$ subregion of $f$ centered at the location $l$ in, e.g., the neighbor of the location $l$. See Fig.~\ref{fig:motivation} and \ref{fig:method} for illustration.

\subsubsection{Semantically Adaptive Kernel Generation}
The SAKG branch aims to generate a semantically adaptive kernel at each location, according to the semantic information.
It consists of four modules, i.e., feature channel compression, semantic kernel generation, feature shuffle, and channel-wise normalization.

\noindent \textbf{Feature Channel Compression.} 
This module is used to reduce the network parameters and computational cost. 
Specifically, the input feature $f$ is fed into a convolutional layer with a $1 {\times} 1$ kernel to compress the input channel from $C$ to $C'$, making SAU have fewer parameters.

\noindent \textbf{Semantic Kernel Generation.} 
This module receives the feature $f_c{\in} \mathbb{R}^{C'{\times} H {\times} W}$ as input (where $H$ and $W$ denote the height and width of the feature map) and tries to generate different semantically adaptive kernels, which can be represented as $f_k{\in} \mathbb{R}^{k^2s^2{\times} H {\times} W}$.
Here $k$ is the semantically adaptive upsampling kernel size and $s$ is the expected upsample scale.
In our experiments, we set $C'{=}64$, $k{=}5$, and $s{=}2$, which achieve good results in most cases.

\begin{figure} [!t] \small
	\centering
	\includegraphics[width=1\linewidth]{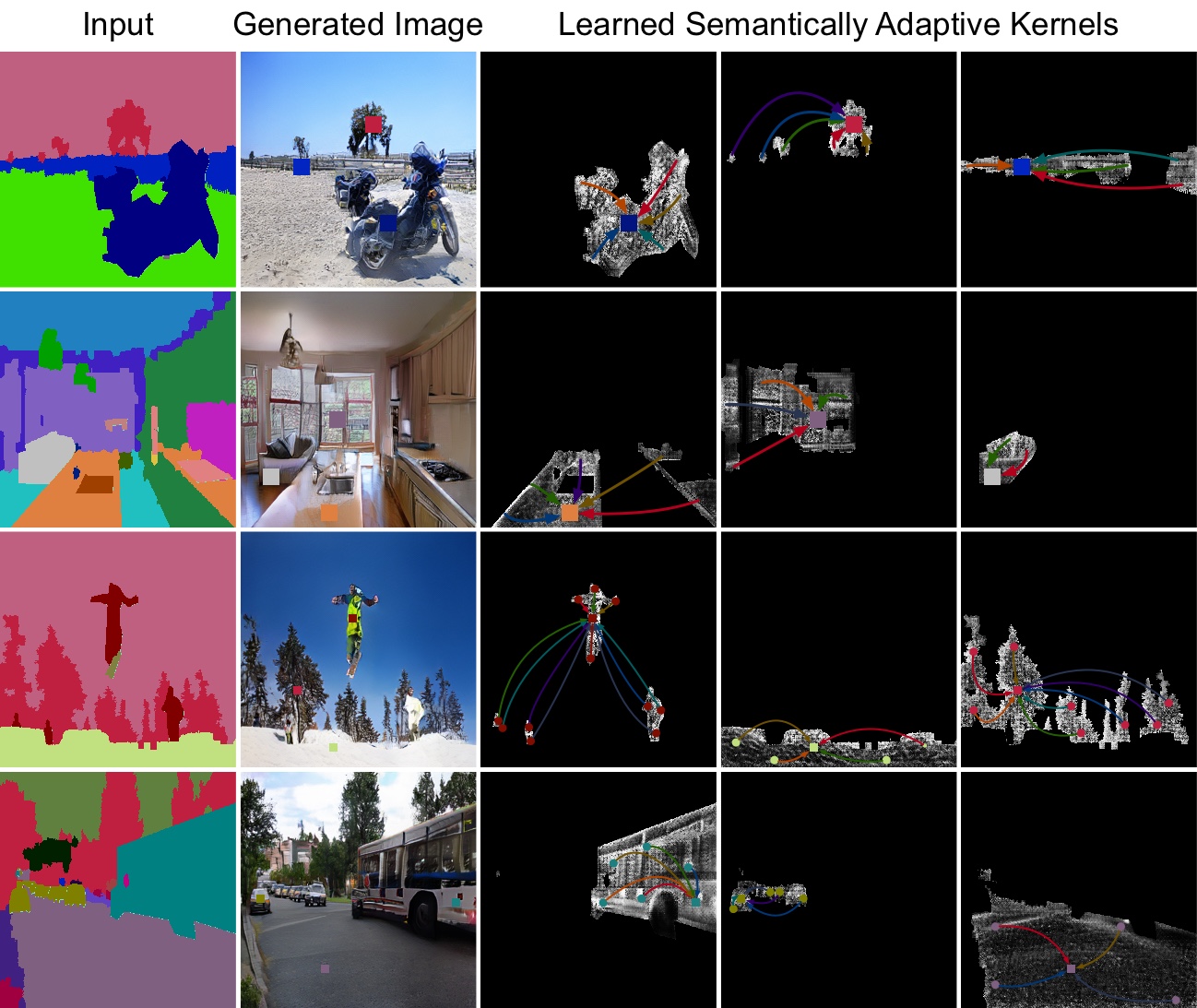}
	\caption{Visualization of semantically adaptive kernels learned on COCO-Stuff.
	In the second column, we show three representative locations in each generated image, with different colored squares. The other three columns show semantically adaptive kernels learned for those three locations, with corresponding color arrows summarizing the most-attended regions for upsampling the target location.
	The network learns to allocate attention according to regions with the same semantic information, rather than just spatial adjacency.}
	\label{fig:attention}
	\vspace{-0.4cm}
\end{figure}

\noindent \textbf{Feature Shuffle.} 
We then feed the feature $f_k$ through a feature shuffle layer to rearrange its elements, leading to a new feature map $f_s{\in} \mathbb{R}^{k^2{\times} Hs {\times} Ws}$, where $k^2{=}k{\times}k$ represents the learned semantic kernel. Note that the learned semantic kernels are quite different at different locations $l'$, as shown in Fig.~\ref{fig:motivation} and \ref{fig:attention}.

\noindent \textbf{Channel-Wise Normalization.} 
Next, we apply a channel-wise softmax operation on each semantic kernel $f_s$ to obtain the normalized kernel $f_n$, i.e., the sum of the weight values in $k^2$ is equal to 1.  
In this way, we can guarantee that information from the combination will not explode.
Moreover, the semantically adaptive kernels can determine which regions to emphasize or suppress according to the semantic information.

\subsubsection{Semantically Adaptive Feature Upsampling}
The SAFU branch aims to upsample the input feature $f$ based on the kernel $f_n$ learned in the SAKG branch, in a semantically adaptive way.
It contains four modules, i.e., feature spatial expansion, sliding local block extraction, feature reshape, and upsampling feature selection.

\noindent \textbf{Feature Spatial Expansion.}
The input feature $f$ is fed into this module to expand the spatial size   from $H{\times}W$ to $Hs{\times} Ws$. 

\noindent \textbf{Sliding Local Block Extraction.} 
Then, the expanded feature $f_e{\in} \mathbb{R}^{C{\times} Hs {\times} Ws}$ is fed into this module to extract a sliding local block of each location in $f_e$, leading to the new feature $f_l{\in} \mathbb{R}^{Ck^2{\times} Hs {\times} Ws}$.

\noindent \textbf{Feature Reshape.}
We next reshape $f_l$ by adding another dimension, resulting in a new feature $f_r{\in} \mathbb{R}^{C {\times} k^2{\times} Hs {\times} Ws}$.
In this way, we can conduct a multiplication between the reshaped local block $f_r$ and the learned kernel $f_n$.

\noindent \textbf{Upsampling Feature Selection.} 
Finally, the feature map $f_r$ and the kernel $f_n$ learned in the SAKG branch are fed into the upsampling feature selection module to generate the final feature map $f'{\in} \mathbb{R}^{C {\times} Hs {\times} Ws}$ in a weighted sum manner.
The computational process at the location $l {=} (i, j)$ can be expressed as follows:
\begin{equation}
\begin{aligned}
f' = \sum_{p=-\lfloor k/2 \rfloor}^{\lfloor k/2 \rfloor} \sum_{q=-\lfloor k/2 \rfloor}^{\lfloor k/2 \rfloor} f_r{(i+p, j+q)} \times f_n{(p, q)}.
\end{aligned}
\end{equation}
In this way, the pixels in the learned kernel $f_n$ contribute to the upsampled pixel $l'$ differently, based on semantic information of features instead of the spatial distance between locations.
The semantics of the upsampled feature map are stronger than those of the original one, since the information from relevant points in a local region can be more attended to, and the pixels with the same semantic label can achieve mutual gains, improving intra-object semantic consistency.

\begin{figure*}[!t] \small
	\centering
\includegraphics[width=0.88\linewidth]{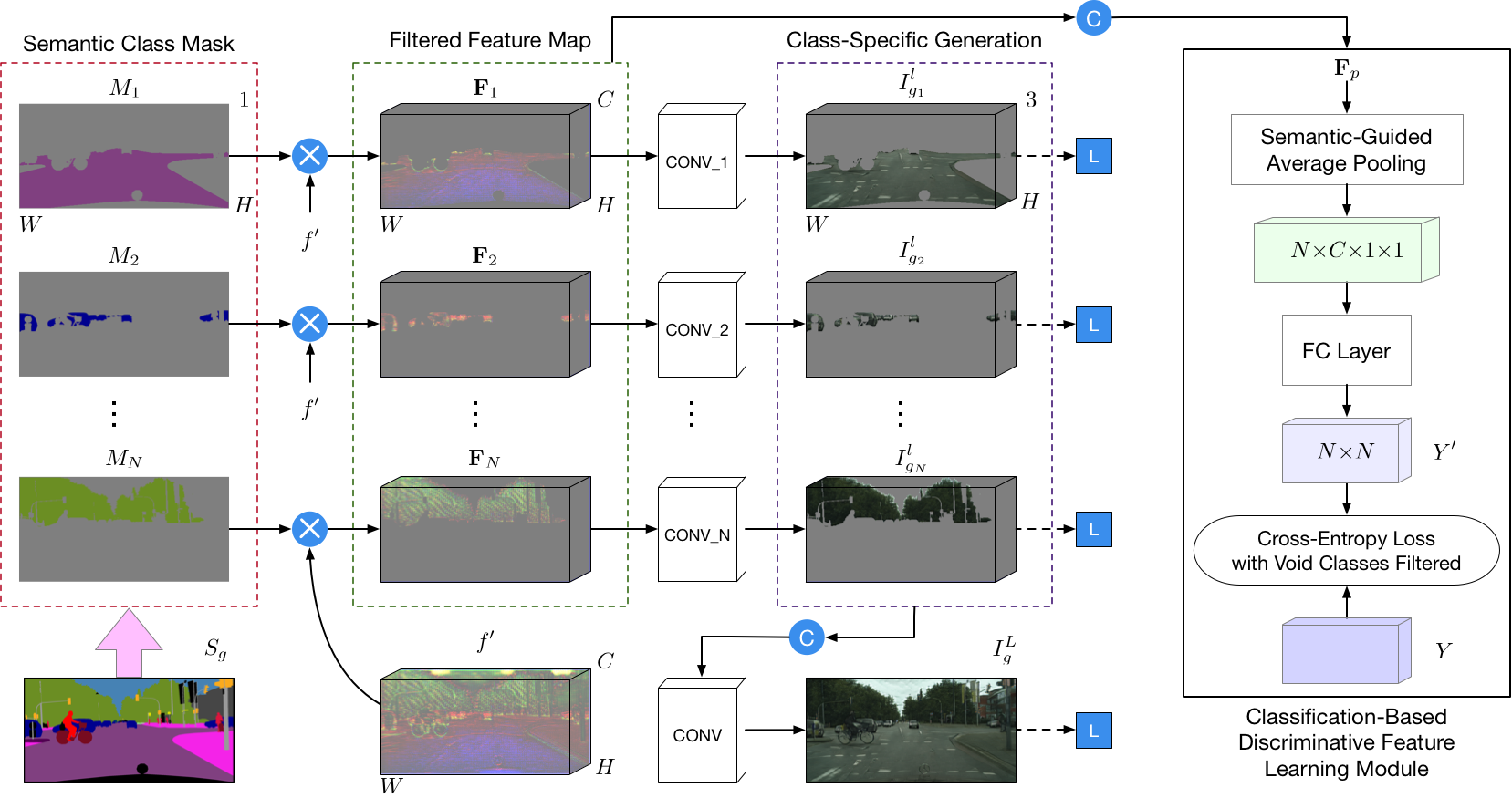}
	\caption{Overview of the proposed local class-specific generator $G_l$, which consists of four steps, i.e., semantic class mask calculation, class-specific feature map filtering, classification-based discriminative feature learning and class-specific generation. A cross-entropy loss with void classes filtered is applied to the feature representation of each class to learn more discriminative class-specific representations. A semantic-mask guided pixel-wise $L1$ loss is applied at the end for class-level reconstruction. The symbols $\otimes$ and $\textcircled{c}$ denote element-wise multiplication and channel-wise concatenation, respectively. Note that we assume the size of $f'$ is $C{\times}H{\times}W$ for simplicity, which is different
    from the one in Fig.~\ref{fig:method} (i.e., $C{\times}Hs{\times}Ws$).}
	\label{fig:local}
	\vspace{-0.4cm}
\end{figure*}

\subsubsection{Why Does the SAU Work Better?}

A short answer is that it can better preserve semantic information compared to other common upsampling methods. 
Specifically, while other methods, such as nearest-neighbor interpolation and deconvolution, are essential parts in almost all state-of-the-art image generation \cite{radford2016unsupervised} and translation \cite{park2019semantic} models, they tend to `pollute' semantic information when performing feature upsampling, since they inevitably incorporate contaminating information from irrelevant regions (see Fig.~\ref{fig:motivation}).

In contrast, the proposed SAU performs feature upsampling using itself, i.e., it uses the pixels belonging to the same semantic label to upsample the feature maps.
Hence, the generator can better preserve semantic information. 
It enjoys the benefit of feature upsampling without losing the input semantic information.
In Fig.~\ref{fig:attention}, we show some examples of the learned semantically adaptive kernels. 
We can easily observe that the proposed SAU achieves upsampling by leveraging complementary features from regions with the same semantic information, rather than local regions of fixed shape to generate consistent objects/scenarios. This further confirms our motivations.

\subsection{Local and Global GAN}

An illustration of the overall framework is shown in Fig.~\ref{fig:framework}. 
The generation module consists of three main parts, i.e., a semantic-guided class-specific generator modeling the local context, an image-level generator modeling the global layout, and a weight-map generator for fusing the local and the global generators. 

\par\noindent\textbf{Class-Specific Local Generation Network.}
As shown in Fig.~\ref{fig:first} and discussed in the introduction, the training data imbalance between different classes and the size difference between semantic objects makes it extremely difficult to generate small object classes and fine details. 
To address this, we propose a novel local class-specific generation network. It constructs a separate generator for each semantic class, thus being able to largely avoid the interference from  large object classes during the joint optimization. 
Each subgeneration branch has independent network parameters and concentrates on a specific class,  therefore being capable of effectively producing similar generation quality for different classes and yielding richer local image details. 

An overview of the local generation network $G_l$ is provided in Fig.~\ref{fig:local}. 
The upsampled feature map $f'$ is multiplied by the semantic mask of each class, i.e., $M_i$, to obtain a filtered class-specific feature map for each one. The mask-guided feature filtering operation can be written as: 
\begin{equation}
\mathrm{\textbf{F}}_i = M_i * f',  \quad i=1,2,...,N,
\label{eqn:fea}
\end{equation}
where $N$ is the number of semantic classes. 
Then, the filtered feature map $\mathrm{\textbf{F}}_i$ is fed into several convolutional layers for the corresponding $i^{th}$ class, which generates an output image $I_{g_i}^l$. To better learn each class, we utilize a semantic mask guided pixel-wise $L1$ reconstruction loss, which can be expressed as follows:
\begin{equation}
\begin{aligned}
\mathcal{L}_{\mathrm{L1}}^{local} = \sum_{i=1}^{N} \mathbb{E}_{I_g, I_{g_i}^l} \lbrack \vert\vert I_g* M_i - I_{g_i}^l \vert\vert_1 \rbrack.
\label{eqn:classl1}
\end{aligned}
\end{equation}
The final output $I_g^L$ from the local generation network can be obtained in two ways. 
The first is by performing an element-wise addition of all the class-specific outputs:
\begin{equation}
I_g^L = I_{g_1}^l \oplus I_{g_2}^l \oplus \cdots \oplus I_{g_N}^l.
\label{eq:local}
\end{equation}
The second is by applying a convolutional operation on all the class-specific outputs, as shown in Fig.~\ref{fig:local}:
\begin{equation}
I_g^L = \mathrm{Conv}(\mathrm{Concat}(I_{g_1}^l, I_{g_2}^l, \cdots, I_{g_N}^l)),
\label{eq:local2}
\end{equation}
where $\mathrm{Concat}(\cdot)$ and $\mathrm{Conv}(\cdot)$ denote a channel-wise concatenation and convolutional operation, respectively.

\begin{figure*}[!t] \small
	\centering
	\includegraphics[width=1\linewidth]{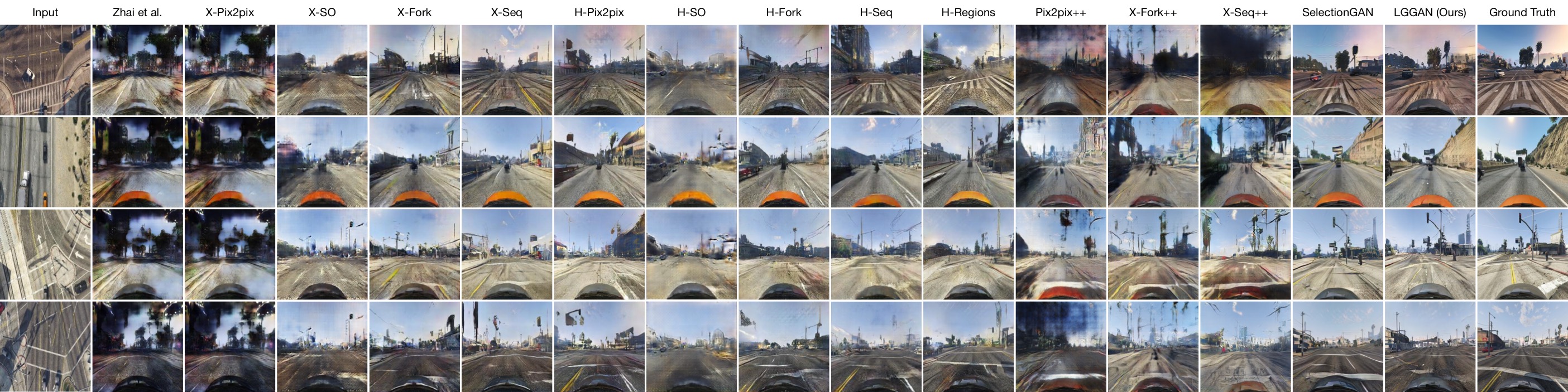}
	\caption{Qualitative comparison of cross-view image translation in a2g direction on SVA.
	}
	\label{fig:sva}
	\vspace{-0.4cm}
\end{figure*}

\noindent \textbf{Class-Specific Discriminative Feature Learning.}
We observe that the filtered feature map $\mathbf{F}_i$ is not able to produce very discriminative class-specific generations, leading to similar generation results for some classes, especially for small-scale object classes.
To provide a more diverse generation for different object classes, we propose a novel classification-based feature learning module to learn more discriminative class-specific feature representations, as shown in Fig.~\ref{fig:local}.
One input sample of the module is a stack of feature maps produced by different local generation branches, i.e.,~$\{\mathbf{F}_1, ..., \mathbf{F}_N\}$.
First, the stacked feature map $\mathbf{F}_p {\in} \mathbb{R}^{N {\times} C {\times} H {\times} W}$ (with $C, H, W$ as the number of feature map channels, height and width, respectively) is fed into a semantic-guided average pooling layer, and we obtain a pooled feature map with dimensions of $N {\times} C {\times} 1 {\times} 1$.
Then the pooled feature map is fed to a fully connected (FC) layer to predict the classification probabilities of the $N$ object classes of the image. 
The output after the FC layer is $Y^{'} {\in} \mathbb{R}^{N {\times} N}$, where $N$ is the number of semantic classes. 
Since, for each filtered feature map $\mathbf{F}_i$ ($i{=}1,...,N$), we predict an $N {\times} 1$ one-hot vector for the probabilities of the $N$ classes.

Since some object classes may not exist in the input semantic mask, the features from the local branches corresponding to the void classes should not contribute to the classification loss. Therefore, we filter the final cross-entropy (CE) loss by multiplying it with a void class indicator for each input sample. The indicator is a one-hot vector $H {=} \{H_i\}_{i=1}^{N}$, with $H_i {=} 1$ for a valid class and $H_i{=}0$ for a void class. Then, the CE loss is defined as follows: 
\begin{equation}
\mathcal{L}_{\mathrm{CE}} = -\sum_{m=1}^{N} H_m\sum_{i=1}^{N}1\{Y(i)=i\} \log(f(\mathbf{F}_i)),
\label{eq:class_loss}
\end{equation}
where $1\{\cdot\}$ is an indicator function having a return of 1 if $Y(i){=}i$, otherwise 0.  $f(\cdot)$ is a classification function which produces a prediction probability given an input feature map $\mathbf{F}_i$. $Y$ is a label set of all object classes.

\par\noindent\textbf{Image-Level Global Generation Network.} Similar to the local generation branch, the upsampled feature map $f'$ is also fed into the global generation subnetwork $G_g$ for image-level generation, as shown in Fig.~\ref{fig:framework}. Global generation is capable of capturing the global structural information or layout of the target images. 
Thus, the global result $I_g^G$ can be obtained through a feed-forward computation: $I_g^G {=} G_g (f').$
Besides the proposed $G_g$, many existing global generators can also be used together with the local generator $G_l$, making the proposed framework very flexible.

\par\noindent\textbf{Pixel-Level Fusion Weight-Map Generation Network.} To better combine the local and global generation subnetworks, we further propose a pixel-level weight map generator $G_w$, which aims at predicting pixel-wise weights to fuse the global generation $I_g^G$ and the local generation $I_g^L$. 
In our implementation, $G_w$ consists of two Transpose Convolution$\rightarrow$InstanceNorm$\rightarrow$ReLU blocks and one Convolution$\rightarrow$InstanceNorm$\rightarrow$ReLU block.
The number of output channels for these three blocks are 128, 64, and~2, respectively. The kernel sizes are $3{\times}3$ with stride~2, except for the last layer which has a kernel size of $1{\times}1$ with stride~1 for dense prediction. 
We predict a two-channel weight map $W_f$ using the following calculation:
\begin{equation}
W_f = \mathrm{Softmax}(G_w(f')),
\label{eqn:final}
\end{equation}
where $\mathrm{Softmax}(\cdot)$ denotes a channel-wise softmax function used for normalization, i.e., the sum of the weight values at the same pixel position is equal to 1. By so doing, we can guarantee that information from the combination will not explode. $W_f$ is sliced to have a weight map $W_g$ for the global branch and a weight map $W_l$ for the local branch. The final fused generation result is calculated as follows:
\begin{equation}
I_g^C = I_g^G \otimes W_g + I_g^L \otimes W_l,
\label{eqn:final_fusion}
\end{equation}
where $\otimes$ is an element-wise multiplication operation.
In this way, the pixel-level weights predicted by $G_w$ directly operate on the output of $G_g$ and $G_l$.
Moreover, the generators $G_w$, $G_g$ and $G_l$ affect and contribute to each other during the model optimization. 

\par\noindent\textbf{Dual-Discriminator.} To exploit prior domain knowledge, i.e., the semantic map, we extend the single-domain vanilla discriminator~\cite{goodfellow2014generative} to a cross-domain structure, which we refer to as the semantic-guided discriminator $D_s$, as shown in Fig.~\ref{fig:framework}. It takes the semantic map $S_g$ and generated image $I_g^C$ (or the real image $I_g$) as inputs: 
\begin{equation}
\begin{aligned}
\mathcal{L}_{\mathrm{CGAN}}(G, D_s) = 
& \mathbb{E}_{S_g, I_g} \left[ \log D_s(S_g, I_g) \right] +  \\
& \mathbb{E}_{S_g, I_g^C} \left[\log (1 - D_s(S_g, I_g^C)) \right],
\end{aligned}
\label{eqn:conditonalgan2}
\end{equation}
aiming to preserve image layout and capture the local information.

For the cross-view image translation task, we also propose another image-guided discriminator $D_i$, which takes the conditional image $I_a$ and the final generated image $I_g^C$ (or the ground-truth image $I_g$) as input:
\begin{equation}
\begin{aligned}
\mathcal{L}_{\mathrm{CGAN}}(G, D_i) = 
& \mathbb{E}_{I_a, I_g} \left[ \log D_i(I_a, I_g) \right] +  \\
& \mathbb{E}_{I_a, I_g^C} \left[\log (1 - D_i(I_a, I_g^C)) \right].
\end{aligned}
\label{eqn:conditonalgan1}
\end{equation}
In this case, the overall loss of our dual-discriminator $D$ is $\mathcal{L}_{\mathrm{CGAN}} {=} \mathcal{L}_{\mathrm{CGAN}}(G, D_i) {+} \mathcal{L}_{\mathrm{CGAN}}(G, D_s)$.

\begin{figure*}[!t] \small
	\centering
	\includegraphics[width=1\linewidth]{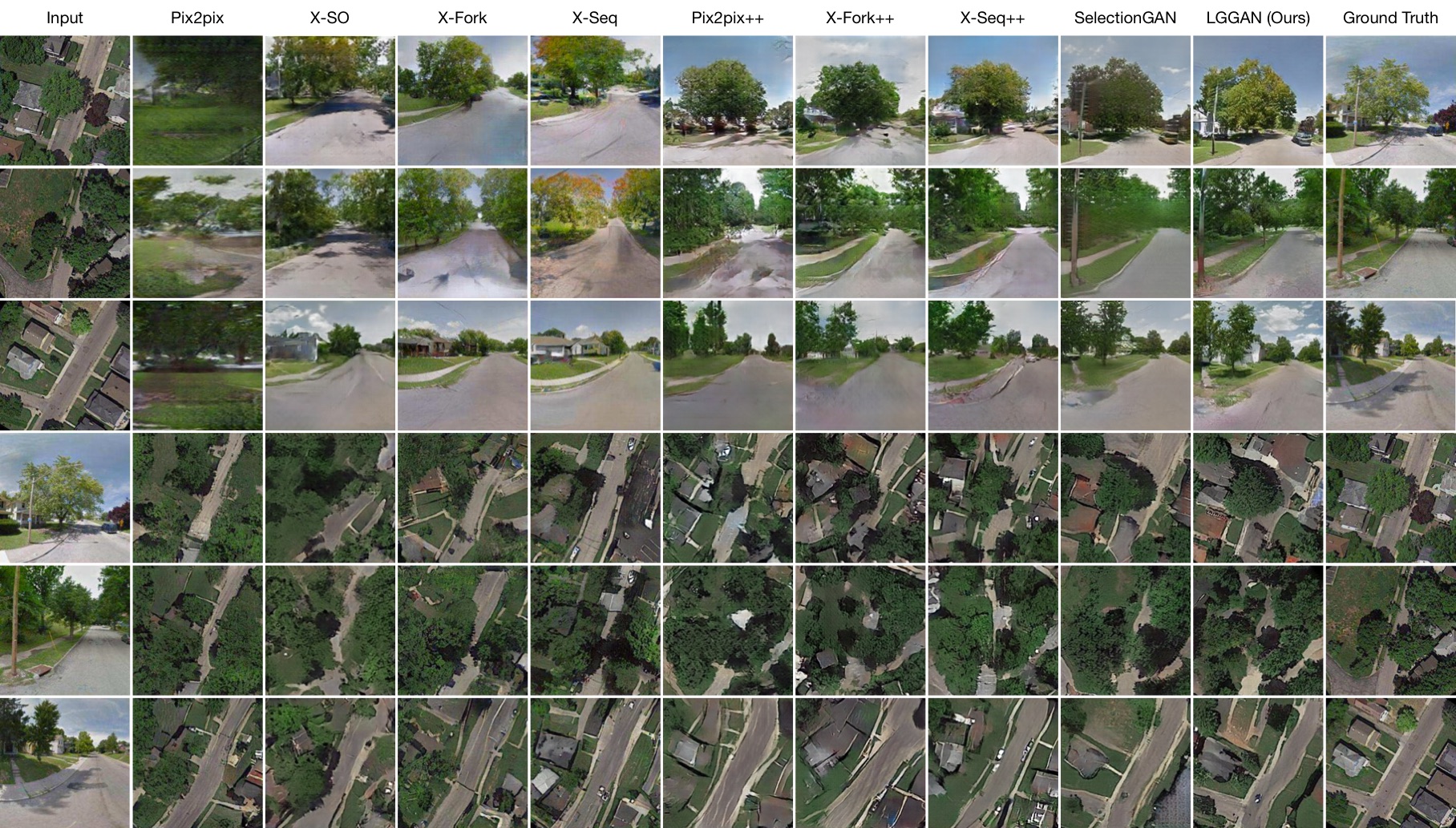}
	\caption{Qualitative comparison of cross-view image translation in both a2g (top three rows) and g2a (bottom three rows) directions on Dayton.
	}
	\label{fig:dayton256}
	\vspace{-0.4cm}
\end{figure*}

\begin{figure*}[!t] \small
	\centering
	\includegraphics[width=1\linewidth]{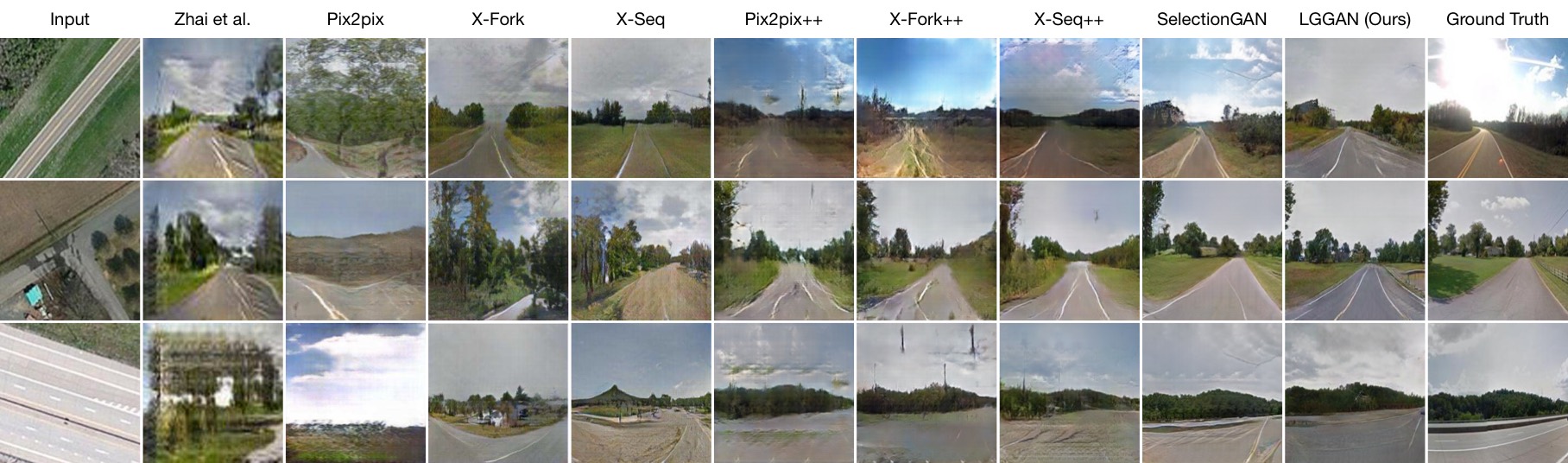}
	\caption{Qualitative comparison of cross-view image translation in a2g direction on CVUSA.
	}
	\label{fig:cvusa}
	\vspace{-0.4cm}
\end{figure*}

%% file: 4Implementation.tex
\section{Experiments}
\label{sec:experiments}

\subsection{Local and Global GAN}

The proposed LGGAN can be applied to different generative tasks, such as cross-view image translation~\cite{tang2019multi} and semantic image synthesis~\cite{park2019semantic}. 

\subsubsection{Results on Cross-View Image Translation}

\noindent \textbf{Datasets and Evaluation Metrics.}
We follow \cite{tang2019multi,regmi2019cross} and perform experiments on the SVA~\cite{palazzi2017learning}, Dayton~\cite{vo2016localizing}, and CVUSA~\cite{workman2015wide} datasets.
Similar to \cite{regmi2019cross,tang2019multi}, we employ inception score (IS), Fr\'echet inception distance (FID)~\cite{heusel2017gans}, accuracy, KL divergence score (KL), structural similarity (SSIM), peak signal-to-noise ratio (PSNR), and sharpness difference (SD) to evaluate the proposed model.

\noindent\textbf{State-of-the-Art Comparisons.}
We compare our LGGAN with several recently proposed state-of-the-art methods, i.e., 
Zhai et al.~\cite{zhai2017predicting},
Pix2pix~\cite{isola2017image}, X-SO~\cite{regmi2019cross}, X-Fork~\cite{regmi2018cross} and X-Seq~\cite{regmi2018cross}.
The comparison results are shown in Tables~\ref{tab:sva}, \ref{tab:dayton}, and~\ref{tab:cvusa}.
We observe that the proposed LGGAN consistently outperforms the competing methods in all metrics. 

To study the effectiveness of our LGGAN, we conduct experiments and compare against using semantic maps and RGB images as input, Pix2pix++~\cite{isola2017image}, X-Fork++~\cite{regmi2018cross}, X-Seq++~\cite{regmi2018cross}, and SelectionGAN~\cite{tang2019multi}.
We implement Pix2pix++, X-Fork++, and X-Seq++ using their public source codes.
The results are shown in Tables~\ref{tab:sva}, \ref{tab:dayton}, and~\ref{tab:cvusa}. 
LGGAN achieves significantly better results than Pix2pix++, X-Fork++, and X-Seq++, confirming its advantage.
A direct comparison with SelectionGAN is also shown in the tables, where our model provides better results on all metrics except those for pixel-level evaluation, i.e., SSIM, PSNR, and SD. 
SelectionGAN achieves slightly better results in these three metrics because it uses a two-stage generation strategy and an attention selection module. 
However, we generate much more photorealistic results than SelectionGAN, as shown in Fig.~\ref{fig:sva}, \ref{fig:dayton256}, and \ref{fig:cvusa}.

\begin{table*}[!t] \small
	\centering
	\caption{Quantitative comparison of cross-view image translation on SVA in the a2g direction. ($\ast$) The inception scores for real (ground truth) data are 3.1282, 2.4932, and  3.4646 for the `all', `top-1', and `top-5' setups, respectively.}
	\resizebox{1\linewidth}{!}{
		\begin{tabular}{lcccccccccccc} \toprule
			\multirow{2}{*}{Method} & \multicolumn{4}{c}{Accuracy (\%) $\uparrow$}& \multicolumn{3}{c}{Inception Score$^\ast$ $\uparrow$} & \multirow{2}{*}{SSIM $\uparrow$} & \multirow{2}{*}{PSNR $\uparrow$} & \multirow{2}{*}{SD $\uparrow$} & \multirow{2}{*}{KL $\downarrow$} & \multirow{2}{*}{FID $\downarrow$} \\ \cmidrule(lr){2-5} \cmidrule(lr){6-8}
			& \multicolumn{2}{c}{Top-1} & \multicolumn{2}{c}{Top-5} & All & Top-1 & Top-5  \\ \midrule
			X-Pix2pix~\cite{isola2017image}&8.5961 &30.3288 &9.0260   &29.9102  & 2.0131 & 1.7221 & 2.2370 & 0.3206 & 17.9944 & 17.0254 & 19.5533 &859.66 \\
			X-SO~\cite{regmi2019cross}     &7.5146 & 30.9507& 10.3905 & 38.9822 & 2.4951 & 1.8940 & 2.6634 &0.4552  & 21.5312 & 17.5285 & 12.0906 &443.79 \\
			X-Fork~\cite{regmi2018cross}   &17.3794&53.4725 &23.8315  &63.5045  & 2.1888 & 1.9776 & 2.3664 & 0.4235 & 21.2400 & 16.9371 & 4.1925  &129.16\\
			X-Seq~\cite{regmi2018cross}    &19.5056&57.1010 &25.8807  &65.3005  &2.2232  & 1.9842 & 2.4344 & 0.4638 & 22.3411 & 17.4138 & 3.7585  &118.70\\ 
			H-Pix2pix~\cite{regmi2019cross}&18.0706& 54.8068& 23.4400 & 62.3072 & 2.1906 & 1.9507 & 2.4069 & 0.4327 & 21.6860 & 16.9468 & 4.2894  &117.13\\
			H-SO~\cite{regmi2019cross}     &5.2444 & 26.4697& 5.2544  & 31.9527 & 2.3202 & 1.9410 & 2.7340 & 0.4457 & 21.7709 & 17.3876 & 12.8761 &1452.88\\
			H-Fork~\cite{regmi2019cross}   &18.0182& 51.0756& 26.6747 & 62.8166 & 2.3202 & 1.9525 & 2.3918 & 0.4240 & 21.6327 & 16.8653 & 4.7246  &109.43\\
			H-Seq~\cite{regmi2019cross}    &20.7391& 57.5378& 28.5517 & 67.4649 & 2.2394 & 1.9892 & 2.4385 & 0.4249 & 21.4770 & 17.5616 & 4.4260  &95.12\\
		H-Regions~\cite{regmi2019cross}&15.4803& 48.0767& 21.8225 & 56.8994 & 2.6328 & 2.0732 & 2.8347 & 0.4044 & 20.9848 & 17.6858 & 6.0638  &88.78\\ \hline		Pix2pix++~\cite{isola2017image}&8.8687 & 34.5434& 9.2713  & 35.7490 & 2.5625 & 2.0879 & 2.7961 & 0.3664 & 17.6549 & 18.4015 & 13.1153 &220.23\\
		X-Fork++~\cite{regmi2018cross} &10.2658& 37.8405& 11.4138 & 38.7976 & 2.4280 & 2.0387 & 2.7630 & 0.3406 & 17.3937 & 18.2153 & 10.1403 &166.33\\
		X-Seq++~\cite{regmi2018cross}  &11.2580& 36.8018& 11.9838 & 36.9231 & 2.6849 & 2.1325 & 2.9397 & 0.3617 & 17.4893 & 18.4122 & 11.8560 &154.80\\
			SelectionGAN \cite{tang2019multi}  &33.9055& 71.8779& 50.8878 & 85.0019 & 2.6576 & 2.1279 & 2.9267 & \textbf{0.5752} & \textbf{24.7136} & \textbf{19.7302} & 2.6183  &26.09\\ 
			LGGAN (Ours) & \textbf{37.0871} &\textbf{75.1314} & \textbf{56.0278} & \textbf{85.4714} & \textbf{2.8088} & \textbf{2.2804} & \textbf{3.1205} & 0.5609 & 24.4779 & 19.6138 & \textbf{2.2922} & \textbf{25.04}\\ \bottomrule
		\end{tabular}}
		\label{tab:sva}
		\vspace{-0.4cm}
	\end{table*}

\begin{table*}[!h] \small
	\centering
	\caption{Quantitative comparison of cross-view image translation on Dayton in the a2g direction.
($\ast$) The inception scores for real (ground truth) data are 3.8319, 2.5753, and 3.9222 for the `all', `top-1', and `top-5' setups, respectively. 
}
		\begin{tabular}{lcccccccccccc} \toprule
			\multirow{2}{*}{Method} & \multicolumn{4}{c}{Accuracy (\%) $\uparrow$} & \multicolumn{3}{c}{Inception Score$^\ast$ $\uparrow$} & \multirow{2}{*}{SSIM $\uparrow$} & \multirow{2}{*}{PSNR $\uparrow$} & \multirow{2}{*}{SD $\uparrow$} & \multirow{2}{*}{KL $\downarrow$}  \\ \cmidrule(lr){2-5} \cmidrule(lr){6-8} 
			& \multicolumn{2}{c}{Top-1} & \multicolumn{2}{c}{Top-5} & All & Top-1 & Top-5 \\ \hline
			Pix2pix \cite{isola2017image}          &6.80 &9.15 &23.55&27.00& 2.8515&1.9342&2.9083 & 0.4180 &17.6291&19.2821& 38.26 $\pm$ 1.88 \\
			X-SO \cite{regmi2019cross} & 27.56 & 41.15 & 57.96 & 73.20 & 2.9459 & 2.0963 & 2.9980 & 0.4772 & 19.6203 & 19.2939 & 7.20 $\pm$ 1.37 \\
			X-Fork \cite{regmi2018cross}           &30.00&48.68&61.57&78.84& 3.0720&2.2402&3.0932 &0.4963&19.8928&19.4533  &6.00 $\pm$ 1.28 \\
			X-Seq \cite{regmi2018cross}               & 30.16&49.85&62.59&80.70& 2.7384&2.1304&2.7674 &0.5031 &20.2803 &19.5258 & 5.93 $\pm$ 1.32 \\ \cmidrule(lr){1-12}
			Pix2pix++~\cite{isola2017image} &32.06 &54.70&63.19&81.01&3.1709&2.1200&3.2001&0.4871&21.6675&18.8504& 5.49 $\pm$ 1.25\\
			X-Fork++~\cite{regmi2018cross} &34.67 & 59.14 &66.37&84.70&3.0737&2.1508&3.0893&0.4982&21.7260&18.9402& 4.59 $\pm$ 1.16 \\
			X-Seq++~\cite{regmi2018cross} & 31.58 & 51.67 &65.21 & 82.48 &3.1703&2.2185&3.2444&0.4912&21.7659&18.9265& 4.94 $\pm$ 1.18 \\
			SelectionGAN~\cite{tang2019multi} & 42.11 & 68.12 & 77.74 & 92.89 & 3.0613 & 2.2707 & 3.1336 & \textbf{0.5938} & \textbf{23.8874} & \textbf{20.0174} & 2.74 $\pm$ 0.86 \\
			LGGAN (Ours) & \textbf{48.17} & \textbf{79.35} & \textbf{81.14} & \textbf{94.91} & \textbf{3.3994} & \textbf{2.3478} & \textbf{3.4261} & 0.5457& 22.9949 & 19.6145 & \textbf{2.18 $\pm$ 0.74}\\
			\bottomrule		
	\end{tabular}
	\label{tab:dayton}
	\vspace{-0.4cm}
\end{table*}

\begin{table*}[!tbp] \small
	\centering
	\caption{Quantitative comparison of cross-view image translation on CVUSA in a2g direction. ($\ast$) The inception scores for real (ground truth) data are 4.8741, 3.2959, and 4.9943 for the `all', `top-1', and `top-5' setups, respectively.}
		\begin{tabular}{lccccccccccccc} \toprule
			\multirow{2}{*}{Method}  & \multicolumn{4}{c}{Accuracy (\%) $\uparrow$}& \multicolumn{3}{c}{Inception Score$^\ast$ $\uparrow$} & \multirow{2}{*}{SSIM $\uparrow$} & \multirow{2}{*}{PSNR $\uparrow$} & \multirow{2}{*}{SD $\uparrow$} & \multirow{2}{*}{KL $\downarrow$} \\ \cmidrule(lr){2-5} \cmidrule(lr){6-8} 
			& \multicolumn{2}{c}{Top-1} & \multicolumn{2}{c}{Top-5} & All & Top-1 & Top-5  \\ \hline
			Zhai et al. \cite{zhai2017predicting}   &13.97 &14.03 &42.09 &52.29 & 1.8434 &1.5171  &1.8666  & 0.4147 &17.4886 &16.6184   & 27.43 $\pm$ 1.63  \\
			Pix2pix \cite{isola2017image} &7.33  &9.25  &25.81 &32.67  & 3.2771 &2.2219 &3.4312  & 0.3923  &17.6578  &18.5239  & 59.81 $\pm$ 2.12   \\ 
			X-SO \cite{regmi2019cross} & 0.29 & 0.21  & 6.14  & 9.08  & 1.7575  & 1.4145  & 1.7791  & 0.3451  & 17.6201  & 16.9919  & 414.25 $\pm$ 2.37 \\
			X-Fork \cite{regmi2018cross}           &20.58 &31.24 &50.51 &63.66  &3.4432 &2.5447 &3.5567  & 0.4356  &19.0509  &18.6706  & 11.71 $\pm$ 1.55 \\
			X-Seq \cite{regmi2018cross}             &15.98 &24.14 &42.91 &54.41  &3.8151 &2.6738 &\textbf{4.0077}  & 0.4231  &18.8067  &18.4378  &15.52 $\pm$ 1.73  \\ \hline
			Pix2pix++ \cite{isola2017image} & 26.45 & 41.87 & 57.26 & 72.87 & 3.2592 & 2.4175 & 3.5078 & 0.4617 & 21.5739 & 18.9044 & 9.47 $\pm$ 1.69 \\
			X-Fork++ \cite{regmi2018cross} &31.03 &49.65 &64.47 &81.16 &3.3758&2.5375&3.5711&0.4769 & 21.6504& 18.9856 &7.18 $\pm$ 1.56\\
			X-Seq++ \cite{regmi2018cross} &34.69&54.61&67.12&83.46&3.3919&2.5474&3.4858 & 0.4740 & 21.6733 & 18.9907 & 5.19 $\pm$ 1.31\\
			SelectionGAN~\cite{tang2019multi} & 41.52  & 65.51  & 74.32  & 89.66 & 3.8074  & 2.7181 & 3.9197  & \textbf{0.5323}  & \textbf{23.1466}  & 19.6100  & 2.96 $\pm$ 0.97  \\ 
			LGGAN (Ours) & \textbf{44.75} & \textbf{70.68} & \textbf{78.76} & \textbf{93.40} & \textbf{3.9180} & \textbf{2.8383} & 3.9878 & 0.5238 & 22.5766 & \textbf{19.7440} & \textbf{2.55 $\pm$ 0.95}  \\            	
			\bottomrule		
		\end{tabular}
		\label{tab:cvusa}
					\vspace{-0.4cm}
	\end{table*}
	
\begin{figure}[!tbp] \small
	\centering
	\includegraphics[width=1\linewidth]{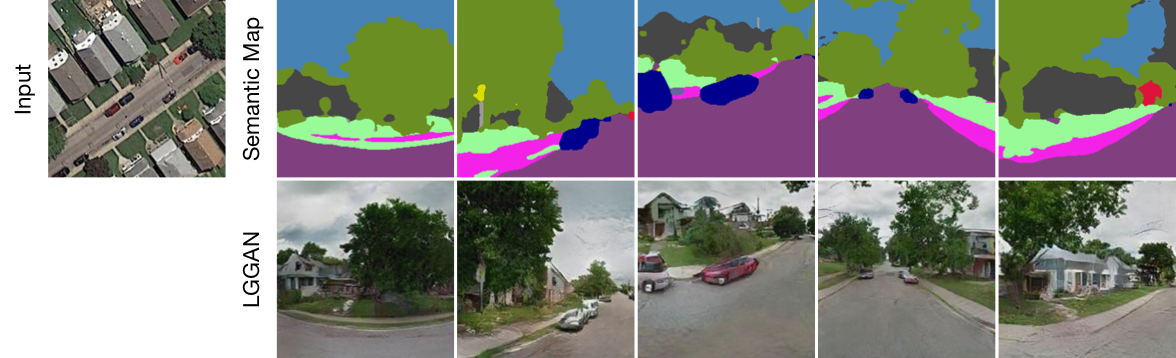}
	\caption{Examples of arbitrary cross-view image translation.}
	\label{fig:arbitrary_results}
	\vspace{-0.4cm}
\end{figure}

\noindent\textbf{Qualitative Evaluation.}
The qualitative results of our model compared with the leading methods are shown in Fig.~\ref{fig:sva}, \ref{fig:dayton256}, and \ref{fig:cvusa}.
The results generated by the proposed LGGAN are visually better than those provided by the existing methods.
Specifically, our method generates clearer details for objects such as cars, buildings, roads, and trees.

\noindent\textbf{Arbitrary Cross-View Image Translation.}
We also follow SelectionGAN \cite{tang2019multi} and show some results on  arbitrary cross-view image translation (Fig.~\ref{fig:arbitrary_results}). We observe that, given an aerial image and a few semantic maps, LGGAN is able to produce the same scene but with different viewpoints.

\noindent\textbf{Network Parameter Comparisons.} 
In Table~\ref{tab:capacity}, we compare the number of network parameters in LGGAN with several state-of-the-art models.
As we can see, the proposed LGGAN achieves superior model capacity and better generation performance compared with existing methods. 

\begin{table}[!tbp] \small
	\centering
	\caption{Comparison of the number of network parameters (M). `G' and `D'  stand for generator and discriminator, respectively.}
	\resizebox{1\linewidth}{!}{%
		\begin{tabular}{cccccc} \toprule
			 Model & Pix2pix \cite{isola2017image}    & X-Fork \cite{regmi2018cross}      & X-Seq  \cite{regmi2018cross}       & SelectionGAN \cite{tang2019multi} & LGGAN \\ \midrule	
			G         & 39.0820  & 39.2163  & 39.0820*2   & 55.4808 & \textbf{12.1913} \\  
			D    & 2.7696    & 2.7696    & 2.7696*2     & 2.7687 & \textbf{2.7678*2}\\ \hline
			Total                  &  41.8516  & 41.9859  & 83.7032   & 58.2495 & \textbf{17.7269} \\        	
			\bottomrule		
	\end{tabular}}
	\vspace{-0.4cm}
	\label{tab:capacity}
\end{table}

\begin{figure*}[t]\small
\centering
\subfigure[GauGAN vs. SMIS vs. LGGAN]{\label{fig:sota_gaugan}\includegraphics[width=0.495\linewidth]{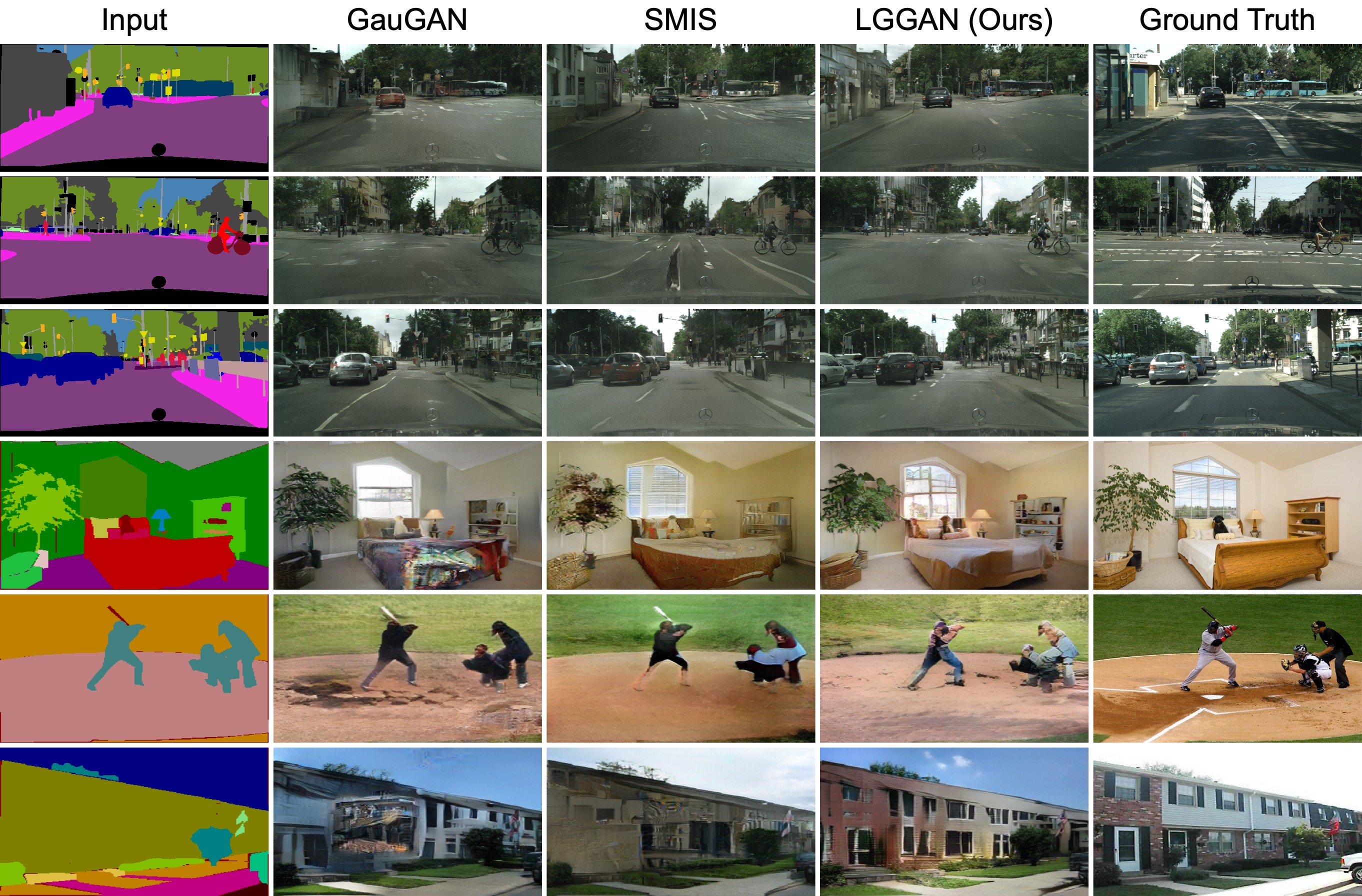}}
\subfigure[SMIS vs. LGGAN vs. LGGAN++ (i.e., LGGAN+SAU)]{\label{fig:sota_city2}\includegraphics[width=0.495\linewidth]{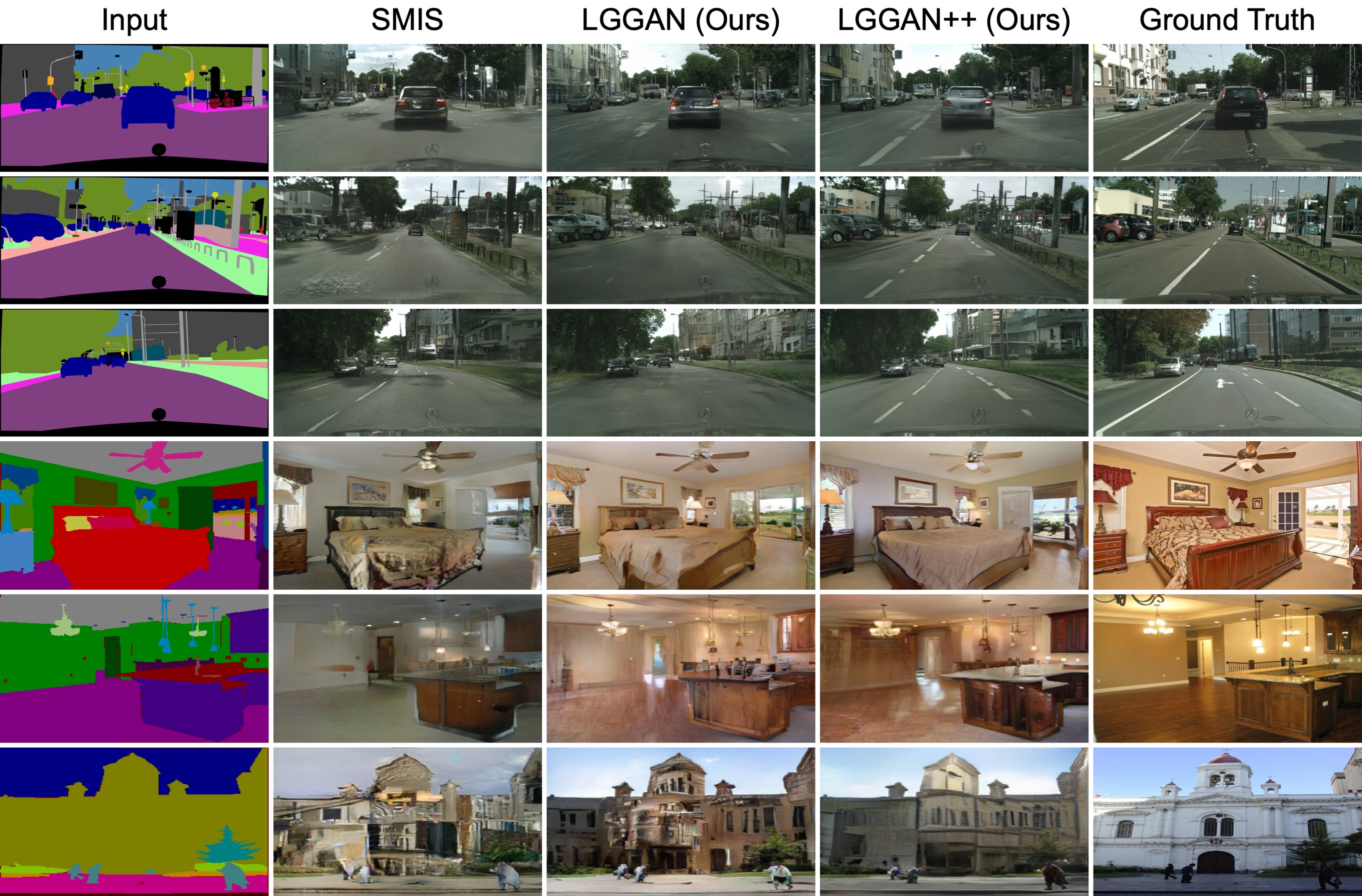}}
\subfigure[SEAN vs. SEAN++ (i.e., SEAN+SAU) vs. LGGAN++]{\label{fig:sota_sean}\includegraphics[width=0.495\linewidth]{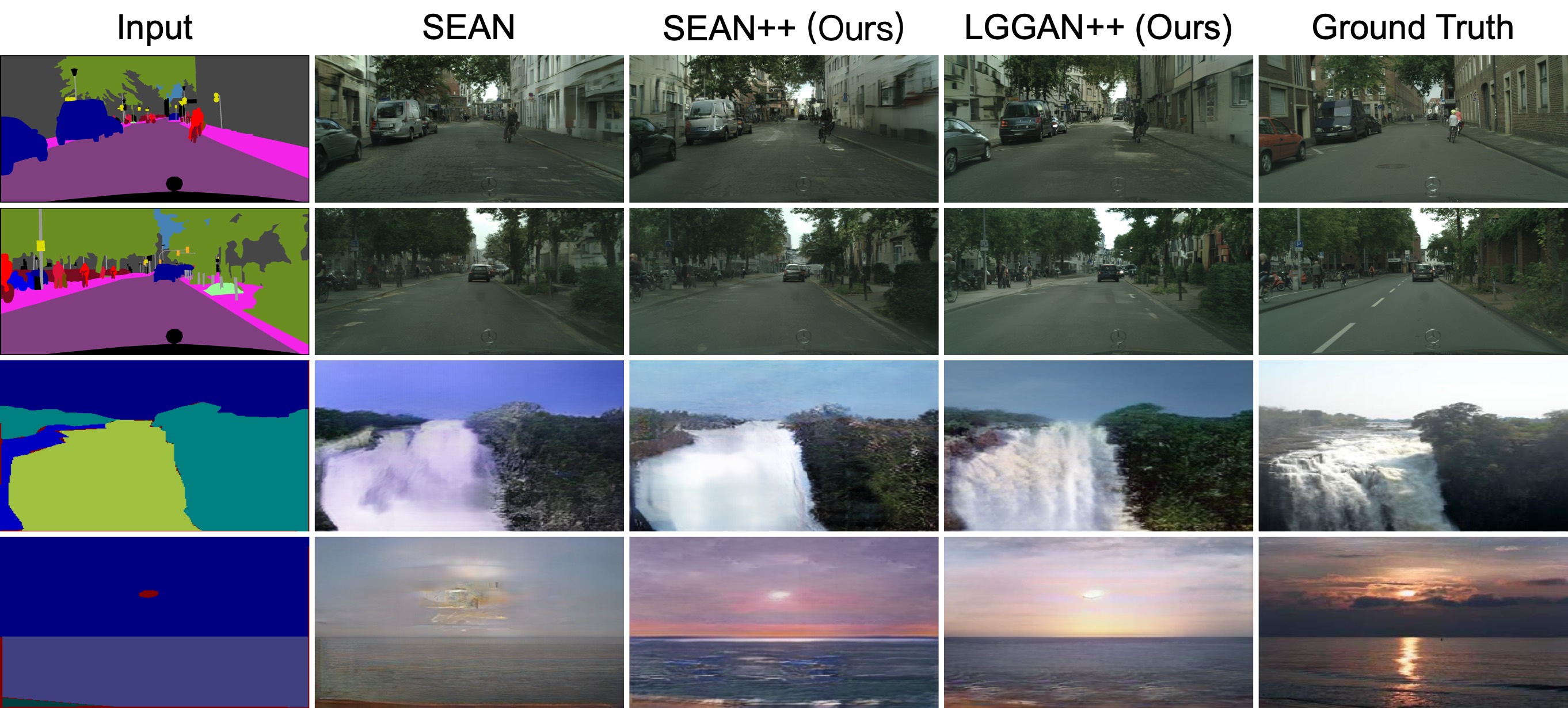}}
\subfigure[MaskGAN vs. MaskGAN++ (i.e., MaskGAN+SAU) vs. LGGAN++]{\label{fig:sota_maskgan}\includegraphics[width=0.495\linewidth]{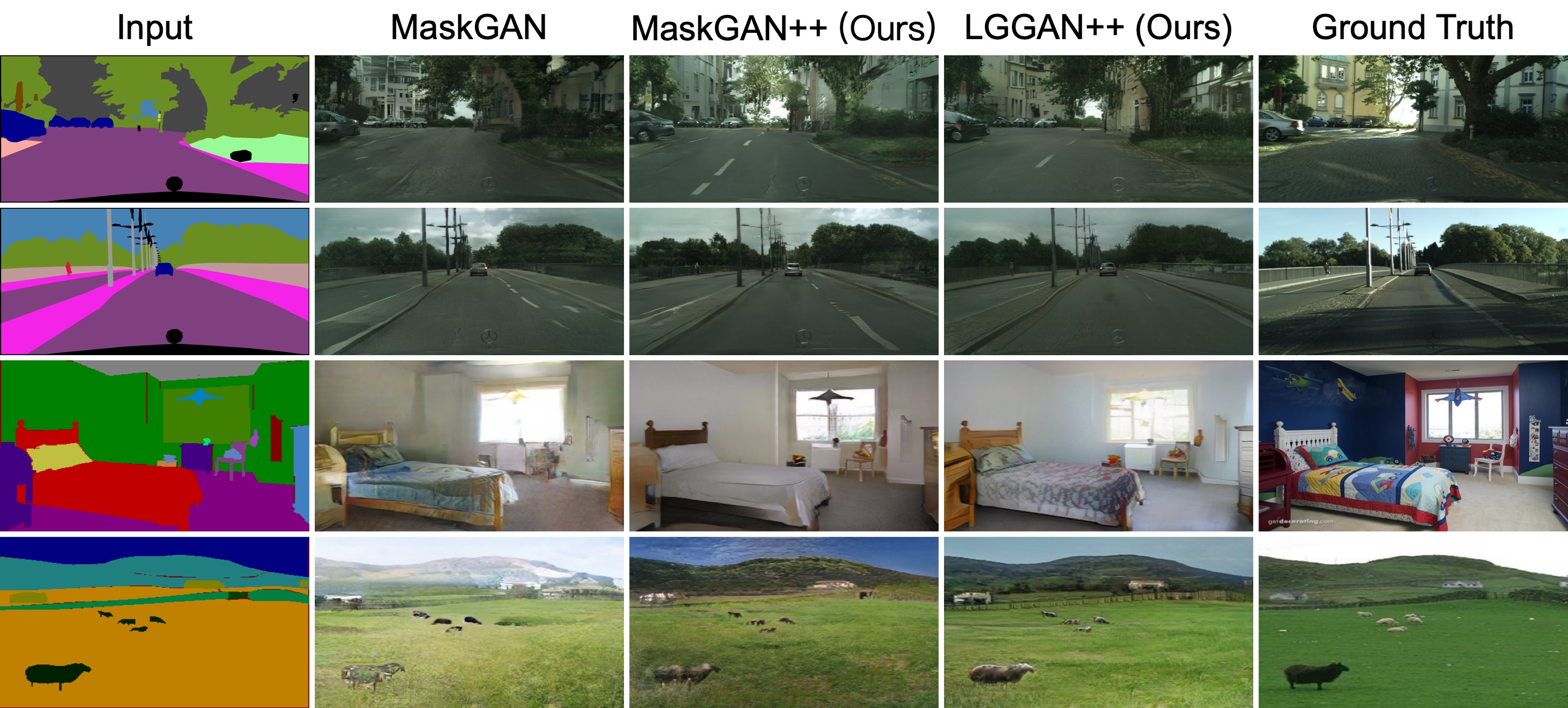}}
\caption{Qualitative comparison of semantic image synthesis on Cityscapes (top) and ADE20K (bottom).
}
\label{fig:cityscapes}
	\vspace{-0.2cm}
\end{figure*}

\begin{table*}[!t] \small
	\centering
		\caption{Quantitative comparison of semantic image synthesis on Cityscapes, ADE20K, and COCO-Stuff.}
		\resizebox{1\linewidth}{!}{%
	\begin{tabular}{llllllllll} \toprule
		\multirow{2}{*}{Method}  & \multicolumn{3}{c}{Cityscapes} & \multicolumn{3}{c}{ADE20K} & \multicolumn{3}{c}{COCO-Stuff} \\ \cmidrule(lr){2-4} \cmidrule(lr){5-7} \cmidrule(lr){8-10} 
		& mIoU $\uparrow$ & Acc $\uparrow$  & FID  $\downarrow$ & mIoU $\uparrow$    & Acc $\uparrow$  & FID  $\downarrow$  & mIoU $\uparrow$    & Acc $\uparrow$  & FID  $\downarrow$ \\ \midrule
		CRN~\cite{chen2017photographic} & 52.4  & 77.1 & 104.7  & 22.4 & 68.8 & 73.3 & 23.7 & 40.4 & 70.4 \\
		SIMS~\cite{qi2018semi}  & 47.2  & 75.5 & 49.7 & -  & - & - & - & - & - \\
		Pix2pixHD~\cite{wang2018high}    & 58.3  & 81.4 & 95.0  & 20.3 & 69.2 & 81.8 & 14.6 & 45.7 & 111.5 \\ 
		SelectionGAN \cite{tang2019multi} & 63.8 & 82.4 & 65.2 & 40.1 & 81.2 & 33.1 & - & - & -\\
		PIS \cite{dundar2020panoptic} & 64.8 & 82.4 & 96.4 & - & - & - & 38.6 & 69.0 & 28.8\\
		TSIT \cite{jiang2020tsit} & 65.9 & 82.7 & 59.2 & 38.6 & 80.8 & 31.6 & - & - & - \\ 
		DAGAN \cite{tang2020dual} & 66.1 & 82.6 & 60.3 & 40.5 & 81.6 & 31.9 & - & - & -\\ 
		SMIS \cite{zhu2020semantically} & 62.3 & 82.1 & 49.8 & 30.4 & 77.1 & 42.2 & - & - & -\\ \hline
		MaskGAN \cite{CelebAMask-HQ} & 63.2 & 81.5 & 65.8 & 35.1 & 76.5 & 33.7 & - & - & - \\
		+ SAU (Ours) & 65.1 (\textbf{+1.9}) & 82.1 (\textbf{+0.6}) & 51.3 (\textbf{-14.5}) & 37.4 (\textbf{+2.3}) & 78.6 (\textbf{+2.1}) & 31.5 (\textbf{-2.2}) & - & - & - \\ \hline
		SEAN \cite{zhu2020sean} & 64.1 & 82.2 & 61.8 & 37.5 & 76.8 & 31.8 & - & - & -\\
		+ SAU (Ours) & 66.3 (\textbf{+2.2}) & 82.7 (\textbf{+0.5}) & 50.3 (\textbf{-11.5}) & 39.2 (\textbf{+1.7}) & 78.9 (\textbf{+2.1}) & 30.1 (\textbf{-1.7}) & - & - & -\\ \hline 
		GauGAN~\cite{park2019semantic} & 62.3  & 81.9 & 71.8  & 38.5 & 79.9 & 33.9 & 37.4 & 67.9 & 22.6 \\
		+ SAU (Ours)  & 65.5 (\textbf{+3.2}) & 82.5 (\textbf{+0.6}) & 48.3 (\textbf{-23.5}) & 39.8 (\textbf{+1.3}) & 80.7 (\textbf{+0.8}) & 32.0 (\textbf{-1.9}) & 39.0 (\textbf{+1.6}) & 69.1 (\textbf{+1.2}) & 20.1 (\textbf{-2.5}) \\ \hline
		LGGAN (Ours) & 68.4 & 83.0 & 57.7 & 41.6 & 81.8 & 31.6 & - & - & -\\ 
		+ SAU (Ours) & 67.7 (-0.7) & 82.9 (-0.1) & 48.1 (\textbf{-9.6}) & 41.4 (-0.2) & 81.5 (-0.3) & 30.5 (\textbf{-1.1}) & - & - & - \\
		\bottomrule
	\end{tabular}}
	\label{tab:sota_semantic}
	\vspace{-0.4cm}
\end{table*}

\subsubsection{Results on Semantic Image Synthesis}
\noindent \textbf{Datasets and Evaluation Metrics}. 
We follow GauGAN~\cite{park2019semantic} and conduct extensive experiments on Cityscapes~\cite{cordts2016cityscapes}, ADE20K~\cite{zhou2017scene}, and COCO-Stuff \cite{caesar2018coco}.
We use mean intersection-over-union (mIoU), pixel accuracy (Acc), and Fr\'echet inception distance (FID)~\cite{heusel2017gans} as evaluation metrics.

\begin{table}[!t]\small
	\centering
	\caption{User study I. The numbers indicate the percentage of users who favor the results of the proposed LGGAN over the competing methods.}
	\begin{tabular}{lcc} \toprule
        AMT $\uparrow$ & Cityscapes                      & ADE20K  \\ \midrule
        LGGAN vs. CRN~\cite{chen2017photographic} & 67.38 & 79.54 \\
        LGGAN vs. Pix2pixHD~\cite{wang2018high} & 56.16   & 85.69 \\ 
        LGGAN vs. SIMS~\cite{qi2018semi} & 54.84          & -   \\
        LGGAN vs. GauGAN~\cite{park2019semantic} & 53.19  & 57.31 \\ 
        LGGAN vs. SMIS \cite{zhu2020semantically} & 56.83 & 62.45\\
        LGGAN vs. MaskGAN \cite{CelebAMask-HQ}    & 54.61 & 60.93 \\
        LGGAN vs. SEAN \cite{zhu2020sean}         & 55.18 & 60.76\\
        LGGAN vs. SelectionGAN~\cite{tang2019multi} & 52.65 & 56.93 \\ \bottomrule
        \end{tabular}
	\label{tab:amt1}
	\vspace{-0.4cm}
\end{table}

\noindent \textbf{State-of-the-Art Comparisons.}
We compare the proposed LGGAN with several leading semantic image synthesis methods, i.e., Pix2pixHD~\cite{wang2018high}, CRN~\cite{chen2017photographic}, SIMS~\cite{qi2018semi}, GauGAN~\cite{park2019semantic}, SelectionGAN \cite{tang2019multi}, TSIT \cite{jiang2020tsit}, PIS \cite{dundar2020panoptic}, DAGAN \cite{tang2020dual}, SMIS~\cite{zhu2020semantically}, MaskGAN~ \cite{CelebAMask-HQ}, and SEAN \cite{zhu2020sean}.
The results in terms of mIoU, Acc, and FID are shown in Table~\ref{tab:sota_semantic}.
The proposed LGGAN outperforms the existing competing methods by a large margin in both mIoU and Acc.
For FID, LGGAN is only worse than SIMS and SMIS on Cityscapes.
However, SIMS and SMIS have poor segmentation performance.
This is because SIMS produces an image by searching and copying image patches from the training dataset. 
The generated images are more realistic since the method uses real image patches.
However, SIMS tends to copy objects with mismatched patches due to the presence queries that cannot be guaranteed to have results.

\begin{figure*}[!t] \small
	\centering
	\includegraphics[width=1\linewidth]{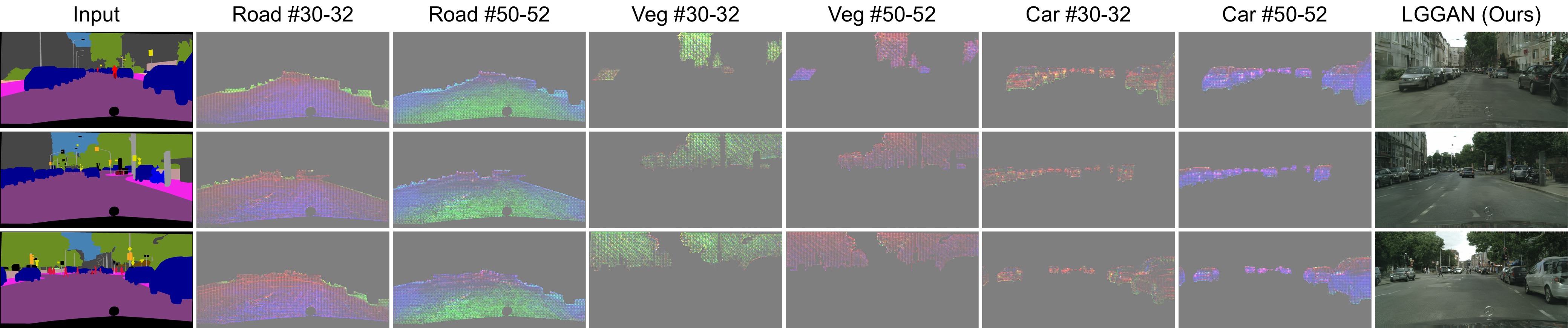}
	\caption{Visualization of class-specific feature maps learned for three different classes, i.e., road, vegetation, and cars. 
	}
	\label{fig:feature}
	\vspace{-0.4cm}
\end{figure*}

\begin{figure*}[!t] \small
	\centering
	\includegraphics[width=1\linewidth]{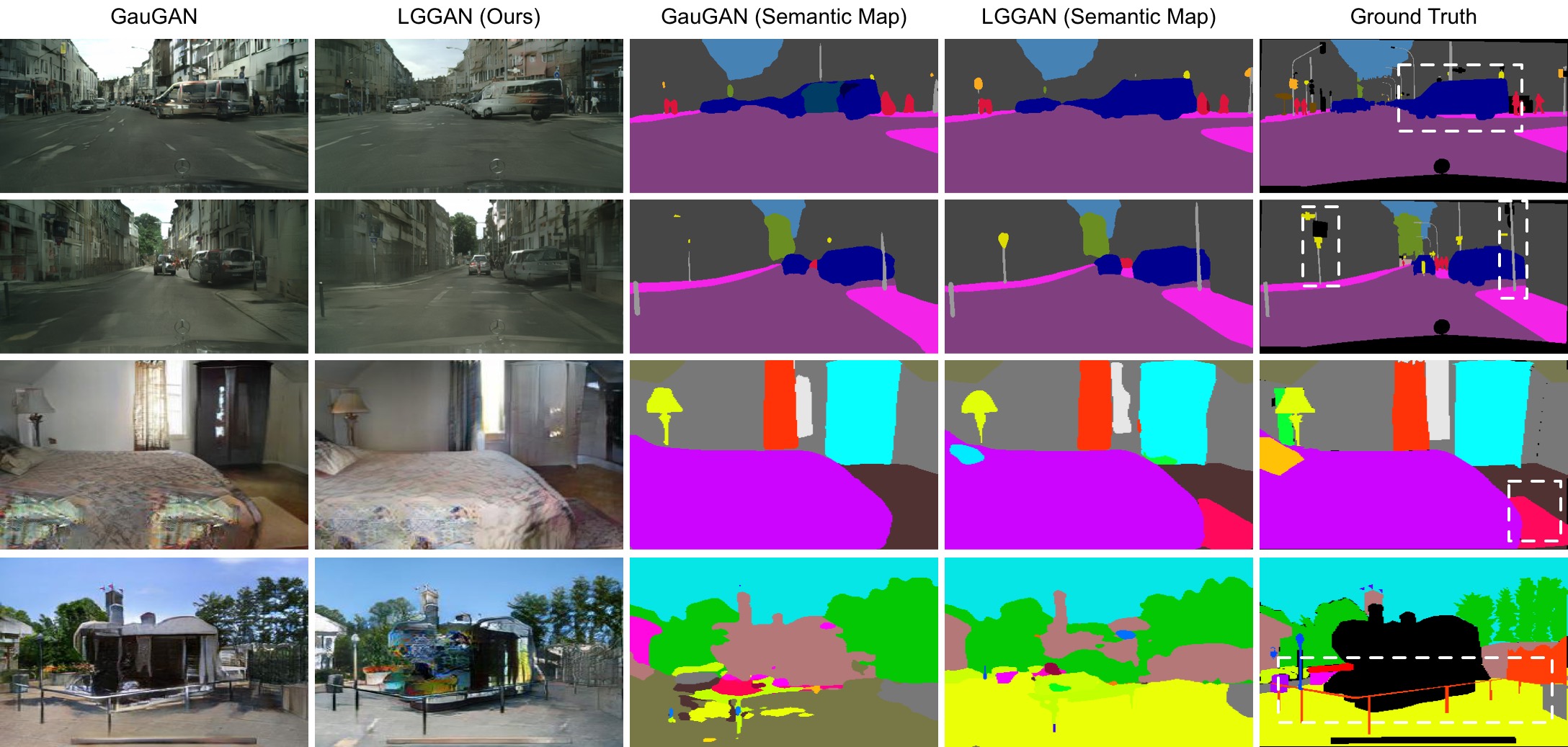}
	\caption{Visualization of semantic maps generated by LGGAN compared with those from GauGAN~\cite{park2019semantic} on Cityscapes (top two rows) and ADE20K (bottom two rows). 
	}
	\label{fig:seg1}
	\vspace{-0.4cm}
\end{figure*}

\noindent \textbf{User Study.}
We follow the evaluation protocol of GauGAN and provide Amazon Mechanical Turk (AMT) results, as shown in Tables~\ref{tab:amt1} and \ref{tab:amt4}.
We see that users favor our synthesized results on both datasets compared with other competing methods.
We also conduct another user study to evaluate the
generated images. Participants are asked to pick the best realistic image among many choices, and we give them unlimited time to respond. Each trial consists of 30 comparisons and is performed by 30 distinct participants. The results are shown in Table \ref{tab:amt3}. We observe that more participants consider our results to be more realistic than those provided by the other methods.

\begin{table}[!t]\small
	\centering
	\caption{User study II. The numbers indicate the percentage of users who favor the results of the proposed LGGAN++ over the competing methods.}
	\begin{tabular}{lcc} \toprule
        AMT $\uparrow$ & Cityscapes                      & ADE20K  \\ \midrule
        LGGAN++ vs. CRN~\cite{chen2017photographic} & 78.17 & 86.31 \\
        LGGAN++ vs. Pix2pixHD~\cite{wang2018high} & 70.25   & 92.86 \\ 
        LGGAN++ vs. SIMS~\cite{qi2018semi} & 68.90          & -   \\
        LGGAN++ vs. GauGAN~\cite{park2019semantic} & 65.57  & 73.36 \\ 
        LGGAN++ vs. SMIS \cite{zhu2020semantically} & 68.87 & 75.90 \\
        LGGAN++ vs. MaskGAN \cite{CelebAMask-HQ}    & 65.32 & 72.17\\
        LGGAN++ vs. SEAN \cite{zhu2020sean}         &66.13 & 76.82\\
        LGGAN++ vs. SelectionGAN~\cite{tang2019multi} & 61.78 & 66.35 \\ 
        LGGAN++ vs. LGGAN &   57.62 & 62.15 \\
        \bottomrule
        \end{tabular}
	\label{tab:amt4}
	\vspace{-0.4cm}
\end{table}

\begin{table}[!t] \small
	\centering
	\caption{User study III. For each comparison, the participant is asked to answer the question, i.e., `Which image is more realistic regardless of the semantic map?'. The numbers indicate the preference percentage of users who favor the results of the corresponding methods.}
		\begin{tabular}{lcc} \toprule
			AMT $\uparrow$ & Cityscapes & ADE20K  \\ \midrule
			CRN~\cite{chen2017photographic}   & 4.6  & 1.2 \\
            Pix2pixHD~\cite{wang2018high}     & 6.1  & 6.6 \\ 
            SIMS~\cite{qi2018semi}            & 7.3  & -   \\
            GauGAN~\cite{park2019semantic}    & 8.2 & 10.1 \\
            SMIS \cite{zhu2020semantically}   & 8.8 & 10.8 \\
            MaskGAN \cite{CelebAMask-HQ}      & 10.5 & 12.3 \\
            SEAN \cite{zhu2020sean}           & 12.1 & 11.5 \\
            SelectionGAN~\cite{tang2019multi} & 10.7 & 13.0 \\ 
            LGGAN (Ours)                      & 14.4 & 16.1 \\ 
            LGGAN++ (Ours)                    & \textbf{17.3} & \textbf{18.4} \\ \bottomrule
	\end{tabular}
	\label{tab:amt3}
	\vspace{-0.4cm}
\end{table}

\noindent \textbf{Qualitative Evaluation.}
The qualitative results compared with the leading methods e.g., GauGAN, SMIS, SEAN, and MaskGAN, are shown in Fig.~\ref{fig:cityscapes}.
We can see that our LGGAN and LGGAN++ generate much better results with fewer visual artifacts than those methods.

\begin{figure*}[t]\small
\centering
\subfigure[Cityscapes]{\label{fig:sota_city}\includegraphics[width=0.654\linewidth]{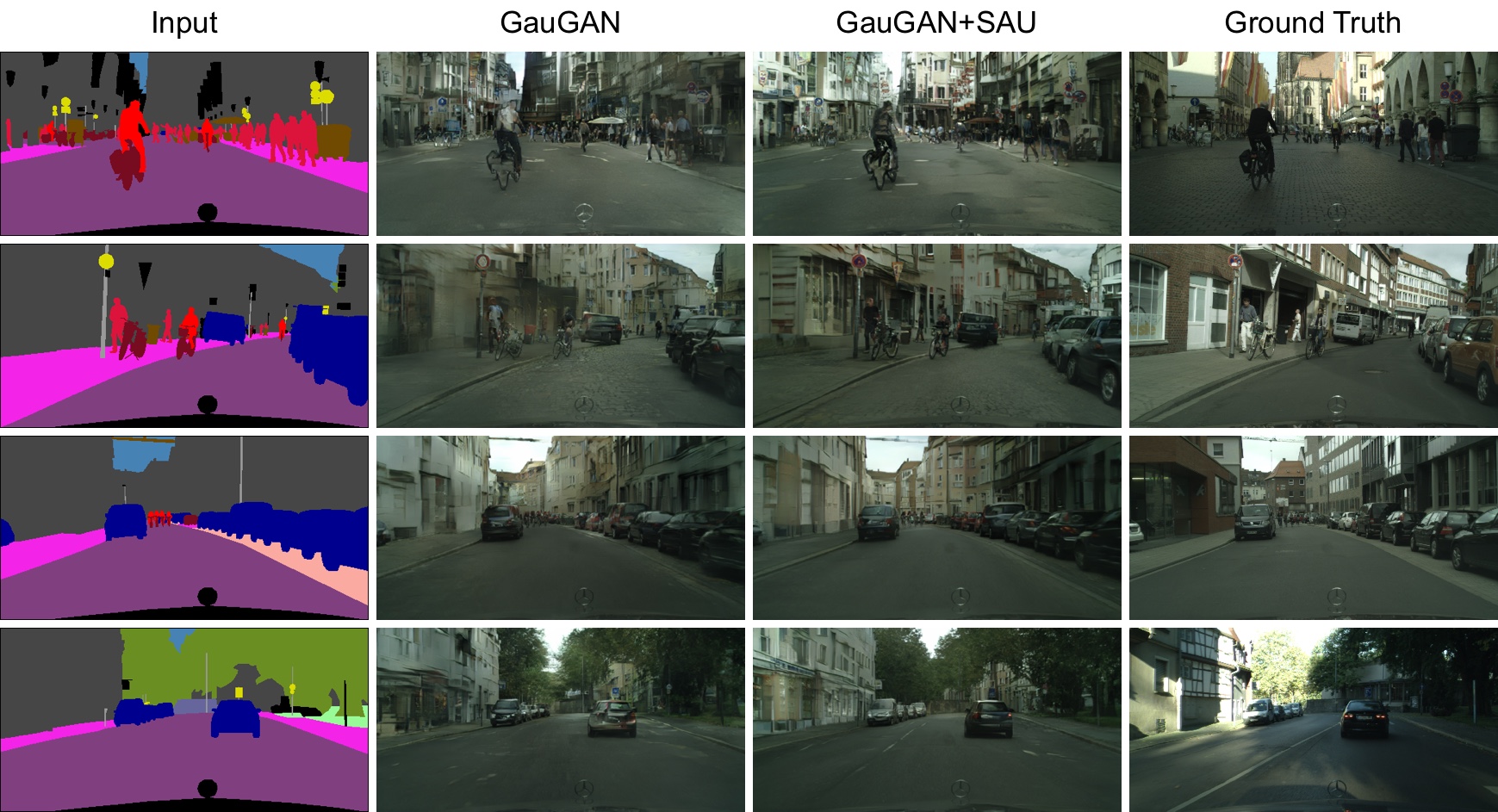}}
\subfigure[Facades]{\label{fig:sota_facades}\includegraphics[width=0.341\linewidth]{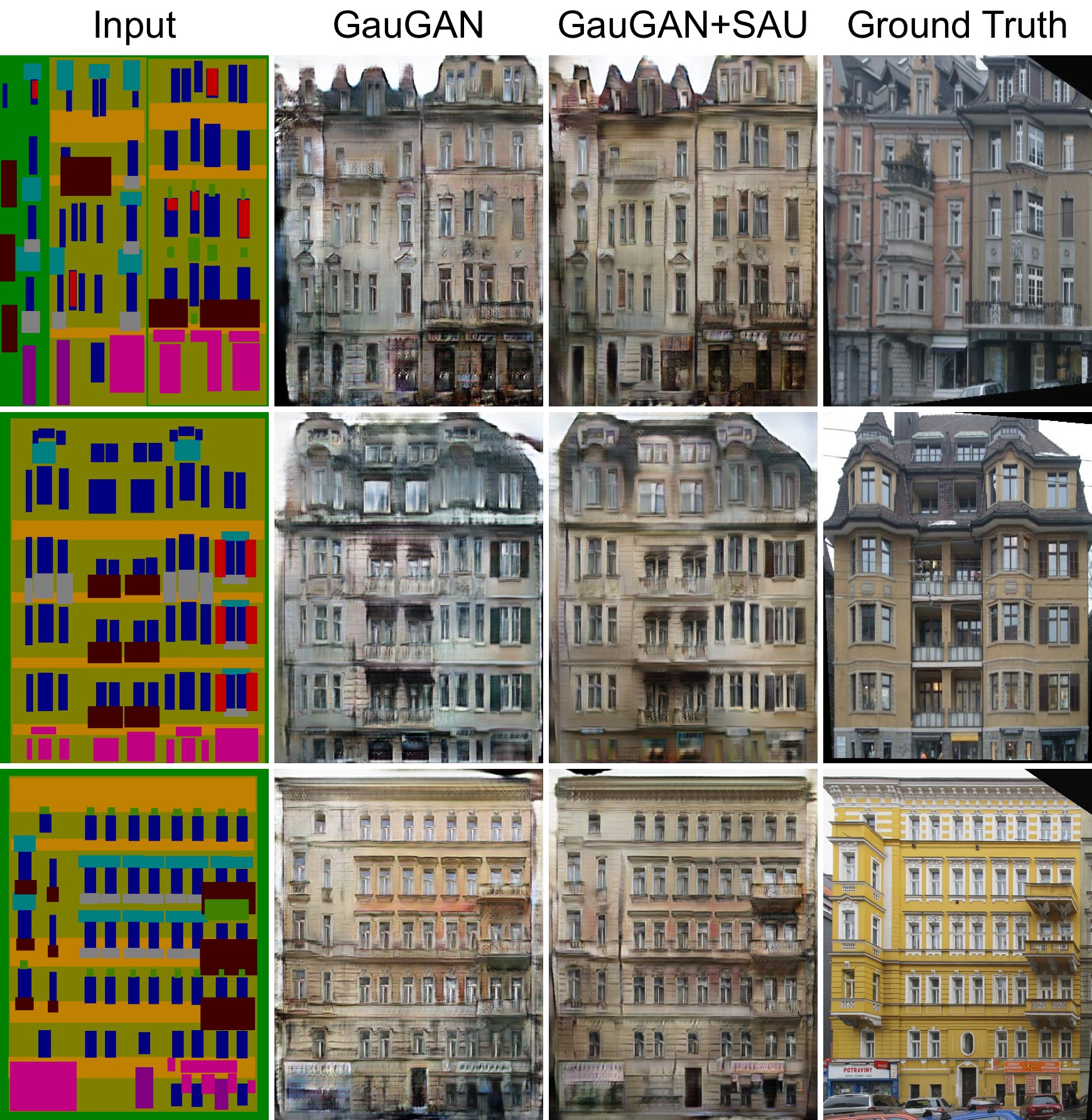}}
\caption{Qualitative comparison of semantic image sythesis on Cityscapes and Facades. 
}
\label{fig:sota_city_facades}
	\vspace{-0.4cm}
\end{figure*}

\begin{figure*}[t]\small
\centering
\subfigure[DeepFashion]{\label{fig:sota_fashion}\includegraphics[width=0.5\linewidth]{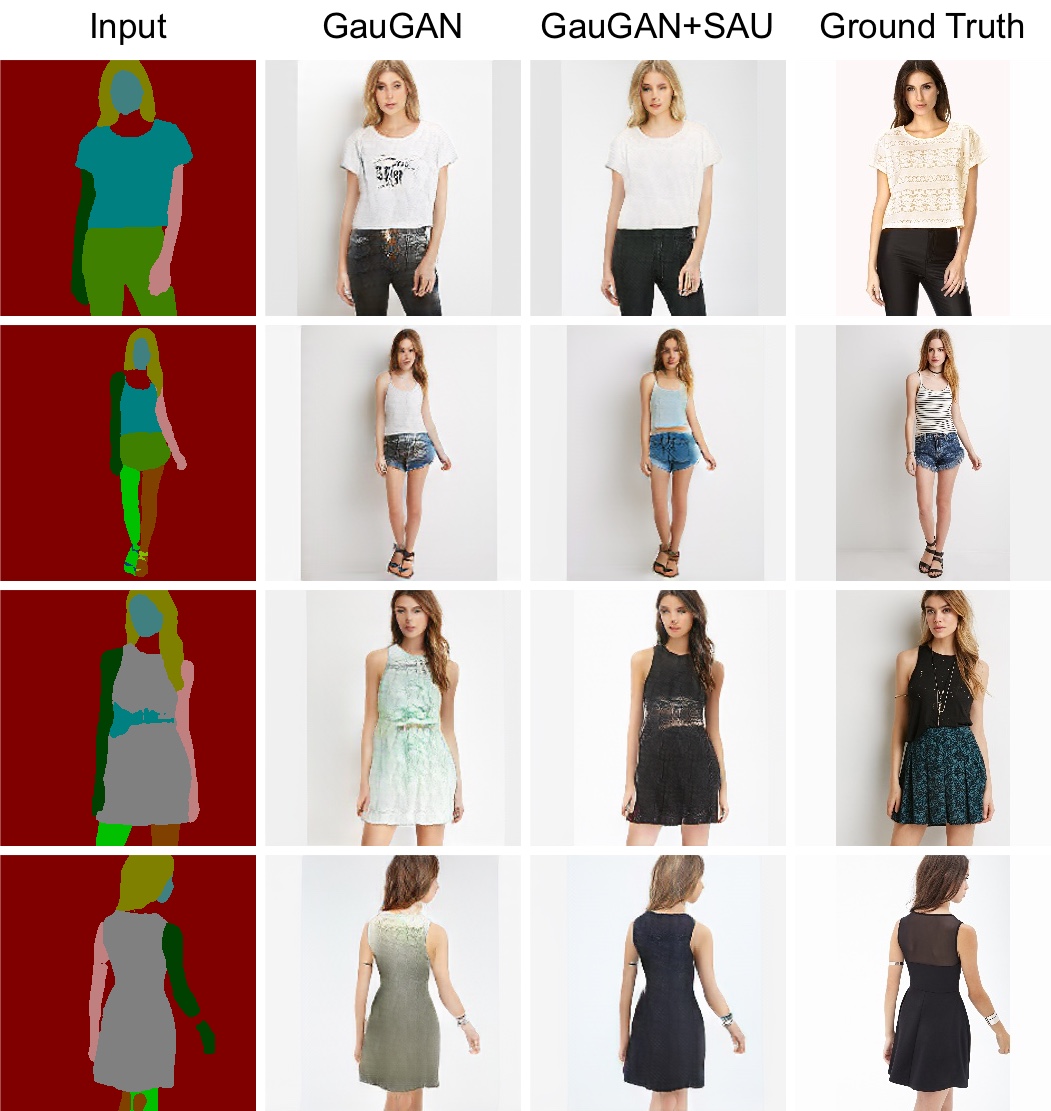}}
\subfigure[CelebAMask-HQ]{\label{fig:sota_celeba}\includegraphics[width=0.488\linewidth]{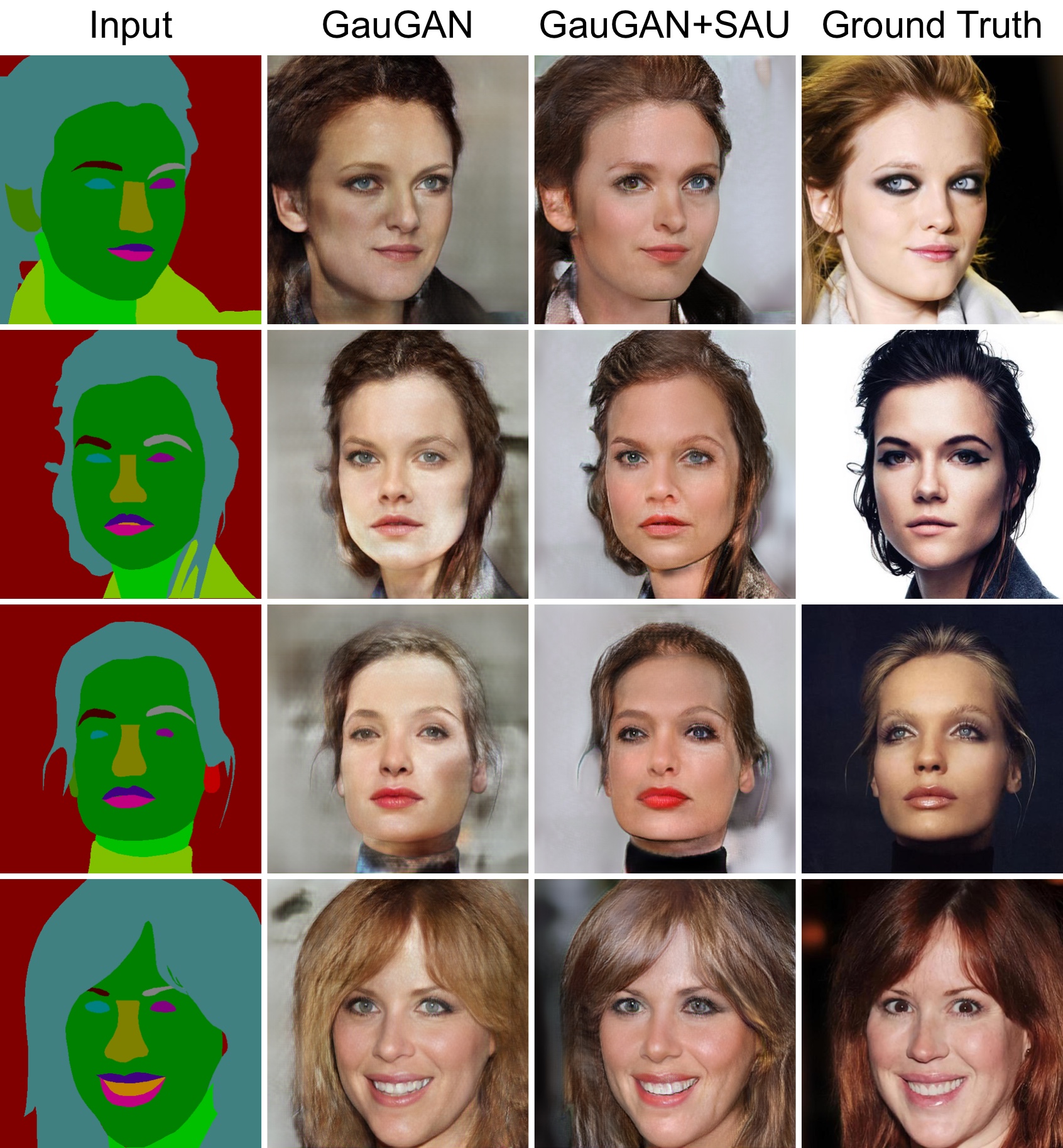}}
\caption{Qualitative comparison of semantic image synthesis on DeepFashion and CelebAMask-HQ.}
\label{fig:sota_fashion_celeba}
	\vspace{-0.4cm}
\end{figure*}

\par\noindent\textbf{Visualization of Learned Feature Maps.}
In Fig.~\ref{fig:feature}, we randomly show some channels from the learned class-specific feature maps (30$^{th}$ to 32$^{th}$, and 50$^{th}$ to 52$^{th}$) on Cityscapes to see if they clearly highlight particular semantic classes.
We show the visualization results for three different classes, i.e., road, vegetation, and cars.
We can easily observe that each local subgenerator effectively learns the deep class-level representations, further confirming our motivations.

\par\noindent \textbf{Visualization of Generated Semantic Maps.}
We follow GauGAN~\cite{park2019semantic} and apply pretrained segmentation networks on the generated images to produce semantic maps, i.e., DRN-D-105~\cite{yu2017dilated} for Cityscapes and UperNet101~\cite{xiao2018unified} for ADE20K.
The generated semantic maps of our LGGAN, GauGAN, and ground truths are shown in Fig.~\ref{fig:seg1}.
As can be seen, LGGAN generates better semantic maps than GauGAN, especially on local textures (`car' in the first row) and small objects (`traffic sign' and `pole' in the second row).

\begin{table}[!t]\small
	\centering
	\caption{Ablation study of the proposed LGGAN on Cityscapes.
	}
        \begin{tabular}{clcc} \toprule
			No. & Setup of LGGAN &  mIoU $\uparrow$ & FID $\downarrow$  \\ \midrule	
			B1 & Ours w/ Global                                         &  62.3   & 71.8  \\
			B2 & B1 + Local (addition)         &  64.6    & 66.1   \\	
			B3 & B1 + Local (convolution)    &  65.8     & 65.6  \\ 
			B4 & B3 + Class Discriminative Loss &  67.0 & 61.3  \\
			B5 & B4 + Weight Map &  \textbf{68.4} & \textbf{57.7} \\ \bottomrule
		\end{tabular}
	\label{tab:results}
	\vspace{-0.4cm}
\end{table}

\begin{figure*}[t]\small
\centering
\subfigure[ADE20K]{\label{fig:sota_ade}\includegraphics[width=0.499\linewidth]{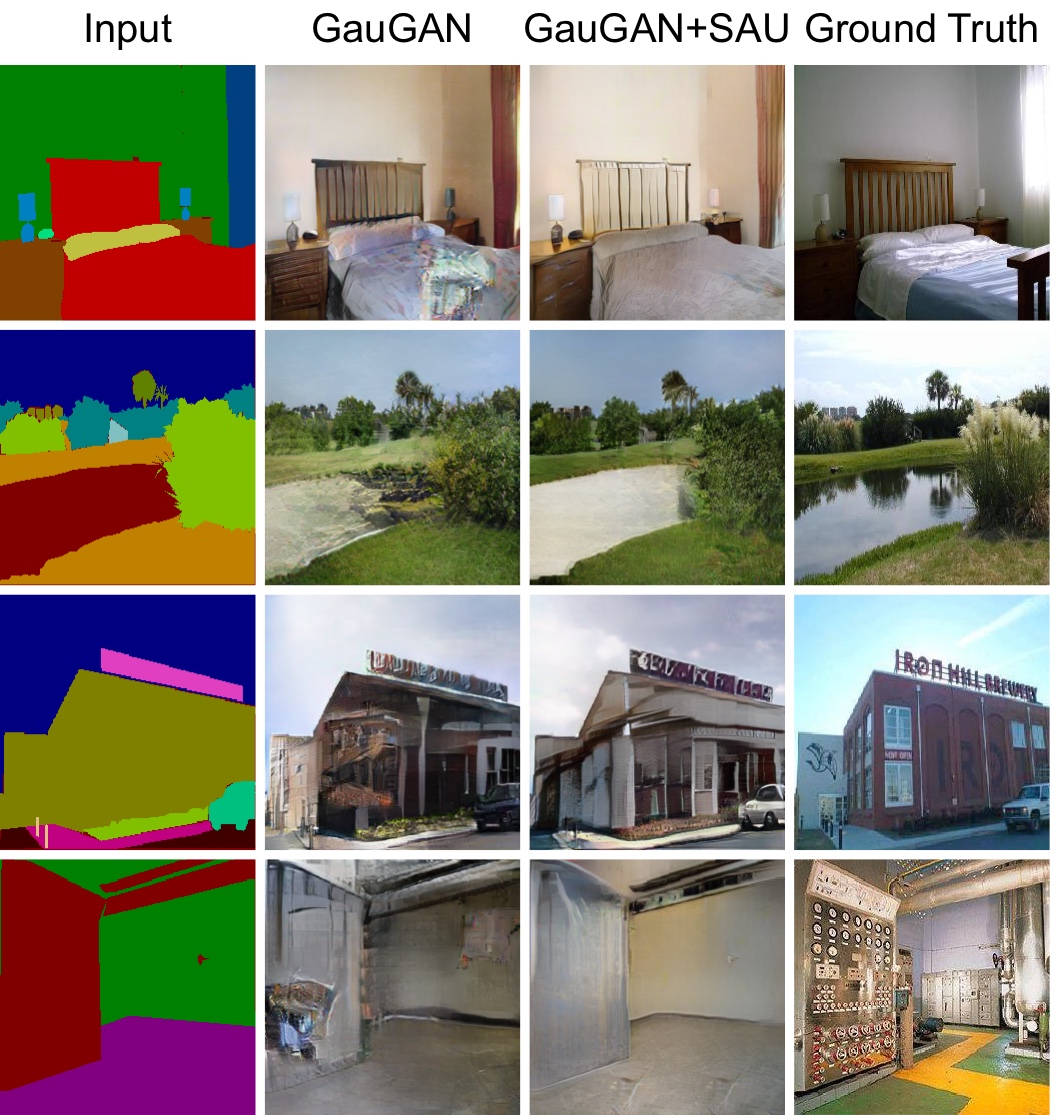}}
\subfigure[COCO-Stuff]{\label{fig:sota_coco}\includegraphics[width=0.4955\linewidth]{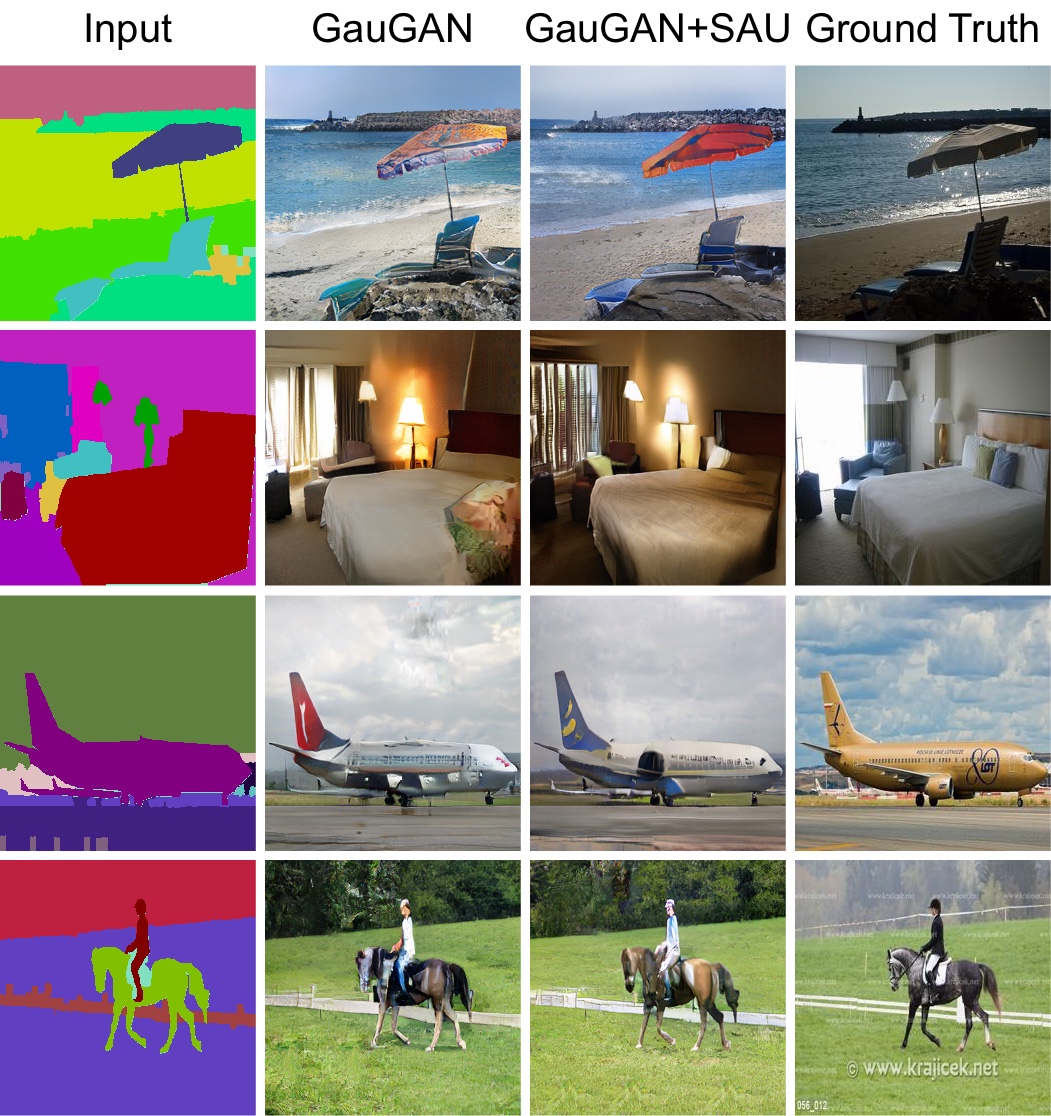}}
\caption{Qualitative comparison of semantic image synthesis on ADE20K and COCO-Stuff.}
\label{fig:sota_ade_coco}
	\vspace{-0.4cm}
\end{figure*}

\begin{figure*}[t]\small
\centering
\subfigure[Cityscapes]{\label{fig:seg}\includegraphics[width=0.749\linewidth]{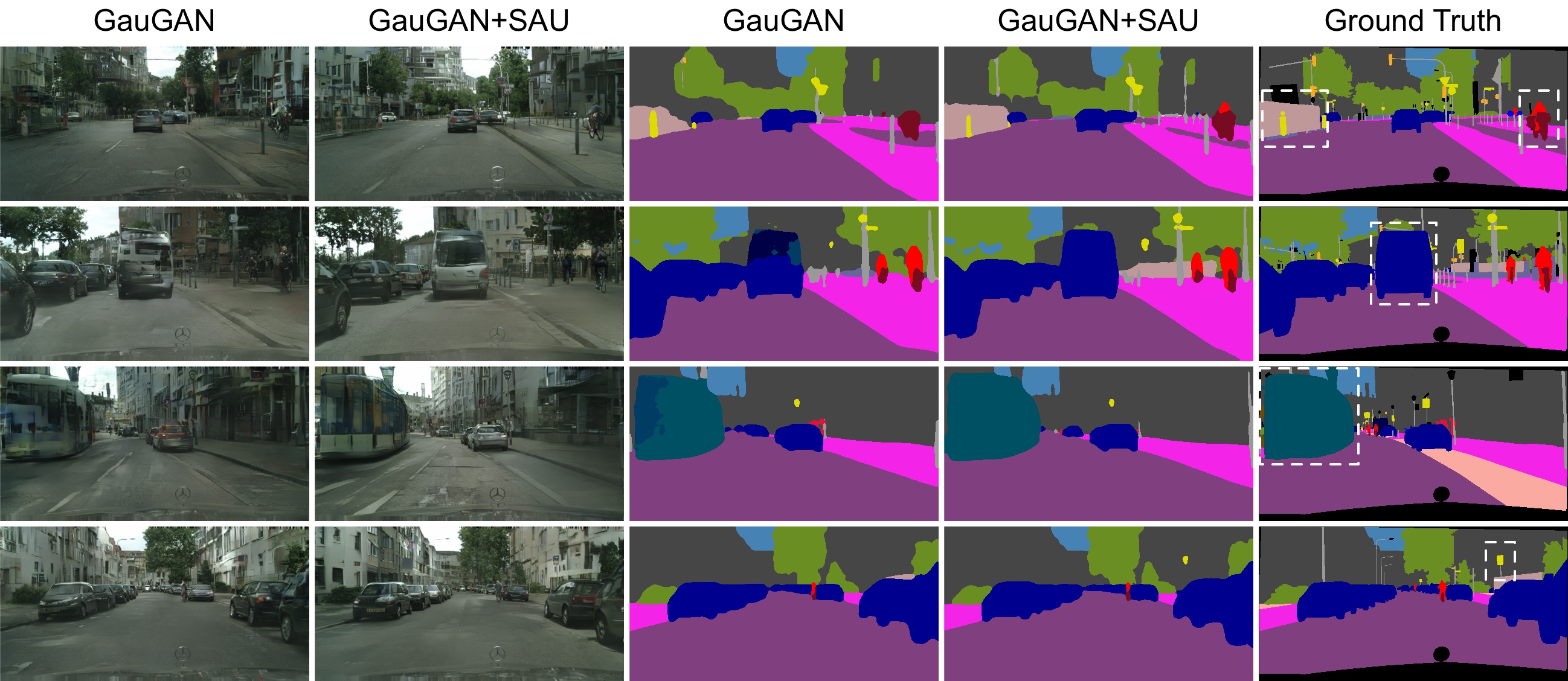}}
\subfigure[CelebAMask-HQ]{\label{fig:super}\includegraphics[width=0.245\linewidth]{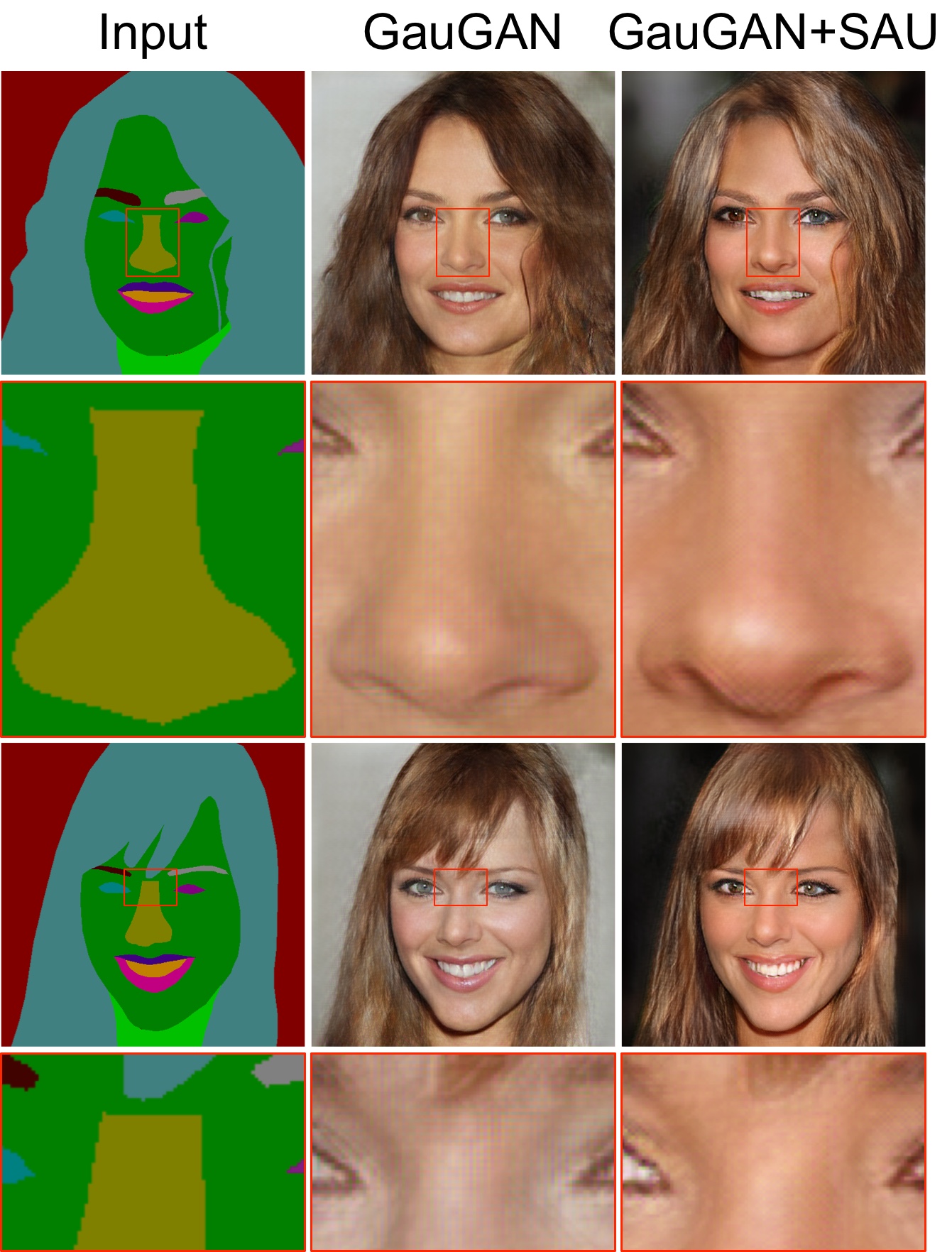}}
\caption{(a) Visualization of semantic maps generated by GauGAN+SAU compared with those from GauGAN~\cite{park2019semantic} on Cityscapes. THe most improved regions are highlighted in the ground truths with white dash boxes. (b) Comparison in a zoomed-in manner on CelebAMask-HQ.}
\label{fig:seg_super}
	\vspace{-0.4cm}
\end{figure*}

\subsubsection{Ablation Study}
\noindent \textbf{Baselines.} We conduct extensive ablation studies on Cityscapes to evaluate different components of our LGGAN.
The proposed LGGAN has five baselines (i.e., B1, B2, B3, B4, B5), as shown in Table~\ref{tab:results}: 
(i) In B1, only the global generator is adopted.
(ii) B2 combines the global generator and the proposed local generator to produce the final results, where the local results are produced using an addition operation, as proposed in Eq.~\eqref{eq:local}.
(iii) B3 is similar to B2 but uses a convolutional layer to generate the local results, as presented in Eq.~\eqref{eq:local2}.
(iv) B4 employs the proposed classification-based discriminative feature learning module.
(v) B5 is our full LGGAN model and adopts the proposed weight map fusion strategy.

\noindent \textbf{Effect of Local and Global Generation.}
The results of the ablation study are shown in Table~\ref{tab:results}.
When using an addition operation to generate the local result, the local and global generation strategy improves the mIoU and FID by 2.3 and 5.7, respectively.
When adopting a convolutional operation to produce the local results, the performance is further improved, with a 3.5 and 6.2 gain in mIoU and FID, respectively.
Both results confirm the effectiveness of the proposed local and global generation framework.
We also provide qualitative results in Fig.~\ref{fig:first}. 
We observe that our full model, i.e., Global+Local,  generates visually better results than both the individual global and local methods.

\noindent \textbf{Effect of Classification-Based Feature Learning.}
B4 significantly outperforms B3, with gains of roughly 1.2 and 4.3 in mIoU and FID, respectively. This means that the model does indeed learn a more discriminative class-specific feature representation, confirming the superiority of our design.

\begin{figure*} [!t] \small
	\centering
	\includegraphics[width=1\linewidth]{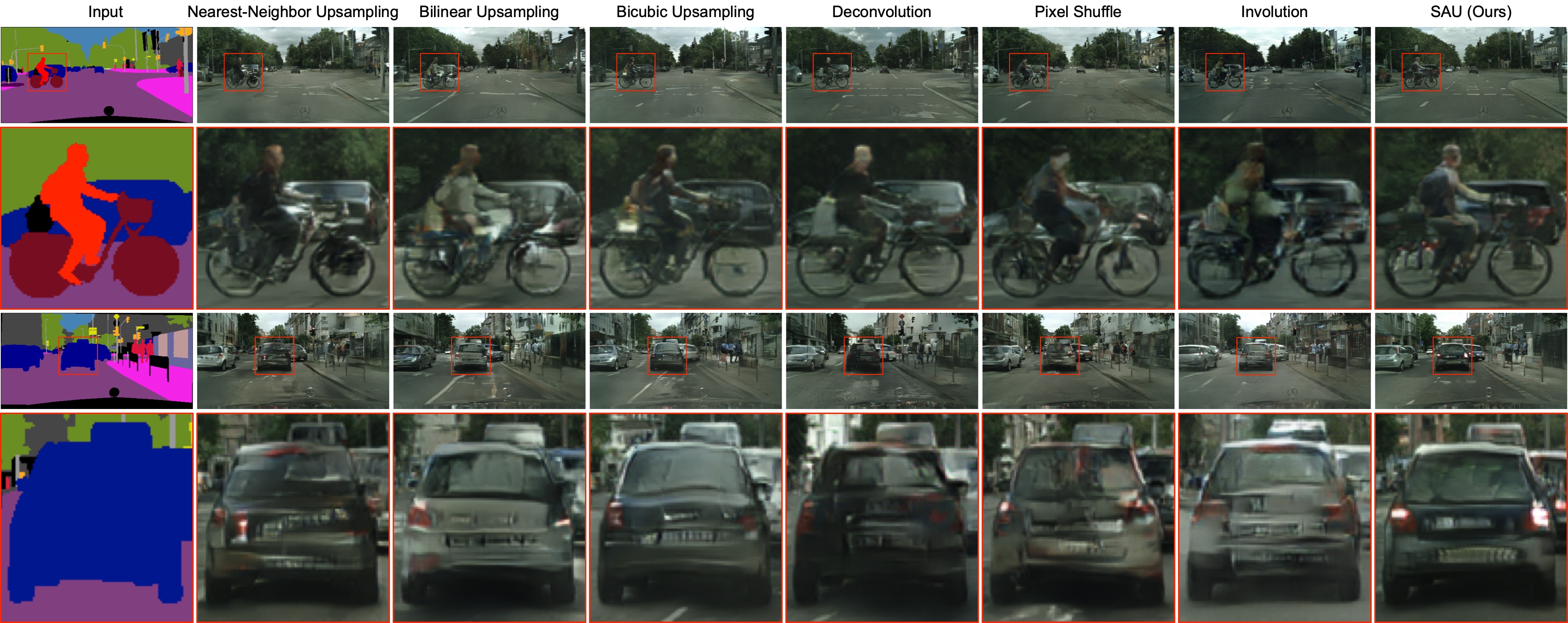}
	\caption{Qualitative comparison of different methods on Cityscapes. Key differences are highlighted by red boxes.}
	\label{fig:abl}
	\vspace{-0.2cm}
\end{figure*}

\begin{table*}[!t] \small
	\centering
	\caption{User study. The numbers indicate the percentage of users who favor the results of the proposed GauGAN+SAU over GauGAN.}
	\begin{tabular}{lcccccc} \toprule
		AMT $\uparrow$ & Cityscapes & ADE20K & COCO-Stuff & DeepFashion & Facades & CelebAMask-HQ \\ \midrule
		GauGAN+SAU (Ours) vs. GauGAN  & 63.8 & 65.7 & 62.4 & 60.1 & 58.3 & 70.5 \\ \bottomrule
	\end{tabular}
	\label{tab:amt}
    \vspace{-0.4cm}
\end{table*}

\noindent \textbf{Effect of Weight Map Fusion.}
By adding the proposed weight map fusion scheme in B5, the overall performance is further boosted, with improvements of 1.4 and 3.6 in mIoU and FID, respectively.
This indicates that the proposed LGGAN can in fact learn complementary information from the local and global generation branches.

\subsection{Semantic-Aware Upsampling}

\noindent \textbf{Datasets.} 
We first follow GauGAN \cite{park2019semantic} and conduct experiments on Cityscapes \cite{cordts2016cityscapes}, ADE20K \cite{zhou2017scene}, and COCO-Stuff \cite{caesar2018coco}.
Then we conduct experiments on three more datasets with diverse scenarios.
1) Facades \cite{tylevcek2013spatial} contains different city images with various architectural styles. The training and test sets have 378 and 228 images, respectively. We resize the images to $512{\times}512$ for high-resolution layout-to-image translation tasks.
2) CelebAMask-HQ \cite{CelebAMask-HQ} contains face images with 19 semantic facial attributes. The training and test sets are made up of 24,183 and 2,842 images, respectively. We also resize these images to $512{\times}512$.
3) DeepFashion \cite{liu2016deepfashion} contains human body images. The number of images in the training and test sets are 30,000 and 2,247, respectively. We resize the images to $256{\times}256$, and use a well-trained model \cite{li2019self} to extract input semantic layouts for this dataset.

\noindent \textbf{Evaluation Metrics.} We follow GauGAN \cite{park2019semantic} and use mIoU, Acc, and FID~\cite{heusel2017gans} as the evaluation metrics on Cityscapes, ADE20K, and COCO-Stuff.
For DeepFashion, CelebAMask-HQ, and Facades, we use FID and Learned Perceptual Image Patch Similarity (LPIPS) \cite{zhang2018unreasonable}.

\subsubsection{Comparisons with State-of-the-Art}

\noindent \textbf{Qualitative Comparisons.}
We first compare the proposed SEAN+SAU and MaskGAN+SAU with SEAN \cite{zhu2020sean} and MaskGAN \cite{CelebAMask-HQ} on both Cityscapes and ADE20K datasets in Fig.~\ref{fig:sota_sean} and \ref{fig:sota_maskgan}. We observe that the models with our SAU generate more realistic reuslts than the models without using it, validating the effectivenes of the proposed SAU.
We then compare the proposed GauGAN+SAU with GauGAN \cite{park2019semantic} on DeepFashion, CelebAMask-HQ, and Facades.
Specifically, we replace the feature upsampling layer in GauGAN with our SAU layer.
Visualization results are shown in Fig.~\ref{fig:sota_facades} and \ref{fig:sota_fashion_celeba}.
We can see that the model with SAU generates more photorealistic results than the original GauGAN.
Moreover, we compare GauGAN and the proposed method in a zoomed-in manner on CelebAMask-HQ in Fig.~\ref{fig:super}. 
As can be seen, the model with our SAU generates more vivid content than the original GauGAN model, further validating the effectiveness of SAU.
Lastly, we compare the proposed method with GauGAN on Cityscapes, ADE20K, and COCO-Stuff.
Comparison results are shown in Fig.~\ref{fig:sota_city} and \ref{fig:sota_ade_coco}.
Our method produces clearer and more visually plausible results than both leading methods, further demonstrating the benefit of our design.

\noindent \textbf{User Study.}
We follow the same evaluation protocol as GauGAN and perform a user study.
The results compared with the original GauGAN are shown in Table~\ref{tab:amt}.
As can be seen, users strongly favor the results generated by our proposed method on all datasets, further validating that the images generated by our upsampling method are more photorealistic.

\begin{table}[!t] \small
	\centering
		\caption{Quantitative comparison of semantic image synthesis on DeepFashion, Facades, and CelebAMask-HQ.}
		\resizebox{1\linewidth}{!}{%
	\begin{tabular}{lcccccc} \toprule
		\multirow{2}{*}{Method}  &  \multicolumn{2}{c}{DeepFashion} & \multicolumn{2}{c}{Facades} & \multicolumn{2}{c}{CelebAMask-HQ}\\ \cmidrule(lr){2-3} \cmidrule(lr){4-5} \cmidrule(lr){6-7} 
		& FID $\downarrow$ & LPIPS $\downarrow$  & FID $\downarrow$ & LPIPS $\downarrow$  & FID $\downarrow$ & LPIPS $\downarrow$ \\ \midrule
		GauGAN~\cite{park2019semantic} & 22.8 & 0.2476 & 116.8 & 0.5437 & 42.2 & 0.4870 \\
		+ SAU (Ours) & \textbf{20.8} & \textbf{0.2446} & \textbf{112.4} & \textbf{0.5387} & \textbf{33.6} & \textbf{0.4788} \\
		\bottomrule
	\end{tabular}}
	\label{tab:sota_semantic2}
	\vspace{-0.4cm}
\end{table}

\noindent \textbf{Quantitative Comparisons.}
Although the user study is most suitable for evaluating the quality of the generated images, we also follow GauGAN and use mIoU, Acc, FID, and LPIPS for quantitative evaluation.
The results compared with several leading methods are shown in Tables \ref{tab:sota_semantic} and~\ref{tab:sota_semantic2}.
Firstly, we observe from Table~\ref{tab:sota_semantic2} that the model with SAU achieves better results than GauGAN on DeepFashion, CelebAMask-HQ, and Facades.
Moreover, from Table~\ref{tab:sota_semantic}, we can see that our methods (i.e., GauGAN+SAU, SEAN+SAU, MaskGAN+SAU) achieve better results compared with the original models on Cityscapes, ADE20K, and COCO-Stuff.

\begin{table*}[!t] \small
	\centering
		\caption{Quantitative comparison of different feature upsampling and enhancement methods on Cityscapes.}
	\begin{tabular}{clcccc} \toprule
		No. & Method  & mIoU $\uparrow$ & Acc $\uparrow$  & FID $\downarrow$ & Params $\downarrow$ \\ \midrule	
		B1 & Nearest-Neighbor Upsampling &62.8 & 81.5 & 58.7 & 93.0M  \\
		B2 & Bilinear Upsampling & 63.9 & 81.9 & 52.9    & 93.0M \\
		B3 & Bicubic Upsampling & 63.7 & 81.8 & 54.4     & 93.0M \\ \hline
		B4 & Deconvolution \cite{noh2015learning} & 63.8 & 82.2& 54.0  & 98.6M\\
		B5 & Pixel Shuffle \cite{shi2016real} & 63.5 & 82.0 & 59.1 & 143.2M \\ 
		B6 & Spatial Attention \cite{fu2019dual} & 64.1 &  81.9 & 56.2 & 97.4M \\ 
		B7 & Involution \cite{li2021involution} & 64.3 & 81.5 & 58.8 & \textbf{93.1}M\\
		B8 & SAU (Ours) &\textbf{65.5 }& \textbf{82.5} & \textbf{48.3} & 93.4M \\  \bottomrule
	\end{tabular}
	\label{tab:abla}
	\vspace{-0.4cm}
\end{table*}

\noindent \textbf{Visualization of Generated Semantic Maps.}
We again follow GauGAN and apply the pretrained DRN-D-105 \cite{yu2017dilated} to the generated Cityscapes images to produce semantic maps.
The results, compared with those produced by GauGAN, are shown in Fig.~\ref{fig:seg}.
We see that the method with our proposed SAU generates more semantically consistent results than the original GauGAN.

\noindent\textbf{LGGAN vs. LGGAN++.}
Our SAU is general and can be seamlessly integrated into existing GANs.
To demonstrate this generalization ability, we conduct more experiments on both Cityscapes and ADE20K.
Specifically, we replace the upsampling layers in LGGAN \cite{tang2020local} with the proposed SAU.
The results are shown in Tables \ref{tab:sota_semantic} and \ref{tab:amt4}, and Fig. \ref{fig:sota_city2}.
We can see from Fig. \ref{fig:sota_city2} that LGGAN++ achieves more photorealistic results than LGGAN, validating the generalization ability of SAU. 
Moreover, LGGAN+SAU (i.e., LGGAN++) achieves a significantly better FID score on both datasets, as shown in Table \ref{tab:sota_semantic}.
Finally, LGGAN++ achieves better user study results than LGGAN, as shown in Table~\ref{tab:amt4}.

\subsubsection{Ablation Study}

\noindent \textbf{Baselines.} 
We conduct an extensive ablation study on Cityscapes to evaluate the effectiveness of the proposed SAU. 
As shown in Table \ref{tab:abla}, B1, B2, and B3 are three traditional upsampling methods. 
B4 and B5 are two learnable upsampling methods.
B6 is the spatial attention module proposed in \cite{fu2019dual}. 
B7 is the involution operation proposed in \cite{li2021involution}.
B8 is our proposed SAU.

\noindent \textbf{Ablation Analysis.}
We first compare the proposed SAU with different upsampling strategies (i.e., B1, B2, B3, B4, B5).
The results of the ablation study are shown in Table \ref{tab:abla} and Fig.~\ref{fig:abl}.
We can see from Table \ref{tab:abla} that the proposed SAU achieves significantly better results than other methods, indicating that the design of effective upsampling methods is critical for this challenging task.
We also observe from Fig.~\ref{fig:abl} that the proposed SAU generates more photorealistic and semantically consistent results with fewer artifacts than other upsampling methods.
Moreover, we add the spatial attention \cite{fu2019dual} and involution \cite{li2021involution} module to GauGAN, obtaining 56.2 and 58.8 in FID, respectively.
We can see that our method still significantly outperforms both spatial attention and involution methods.

\noindent \textbf{Model Parameter Comparisons.}
We also compare the number of generator parameters with different baselines.
The results are shown in Table~\ref{tab:abla}. 
Traditional upsampling methods (B1, B2, and B3) have the same number of parameters. 
Moreover, we can see that the proposed SAU achieves superior model capacity compared to the learnable upsampling methods (i.e., B4 and B5) and spatial attention (i.e., B6).

%% file: 5Conclusion.tex
\section{Conclusion}
We propose a local class-specific and global image-level generative adversarial network (LGGAN) for semantic-guided image generation. 
The proposed LGGAN contains three generation branches, i.e., global image-level generation, local class-level generation, and pixel-level fusion weight map generation, respectively.
A new class-specific local generation network is designed to alleviate the influence of imbalanced training data and the size difference of  objects for joint learning. 
To learn more discriminative and class-specific feature representations, a novel classification module is further proposed. 
Moreover, we introduce a novel semantic-aware upsampling method, which is able to aggregate semantic information in the input layout and adaptively conduct class-specific upsampling during the translation process.
Experimental results demonstrate the superiority of the proposed approach, which achieves the new state-of-the-art on both cross-view image translation and semantic image synthesis, on nine public datasets.

\noindent \textbf{Acknowledgments.}
This work was partially supported by the EU H2020 AI4Media Project under Grant 951911, the PRIN project CREATIVE Prot. 2020ZSL9F9, and the National Natural Science Foundation of China (No. 61929104).

%% file: 6appendices.tex
\appendices

\section{Applications}

\begin{figure*}[!t]\small
\centering
\subfigure[Cityscapes]{\label{fig:app_city}\includegraphics[width=0.615\linewidth]{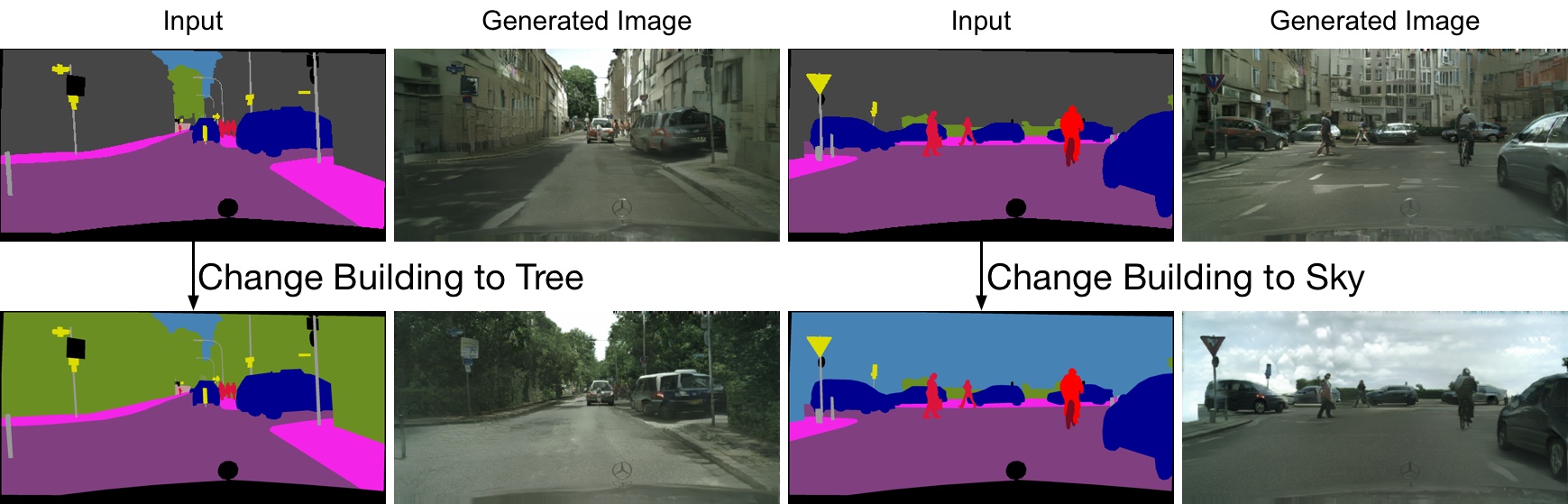}}
\subfigure[ADE20K]{\label{fig:app_ade}\includegraphics[width=0.185\linewidth]{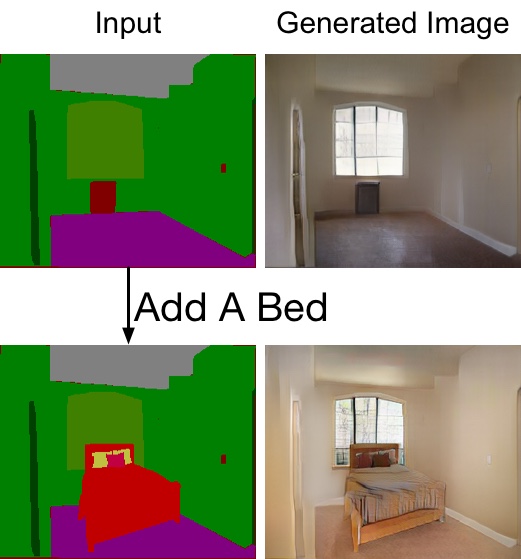}}
\subfigure[COCO-Stuff]{\label{fig:app_coco}\includegraphics[width=0.185\linewidth]{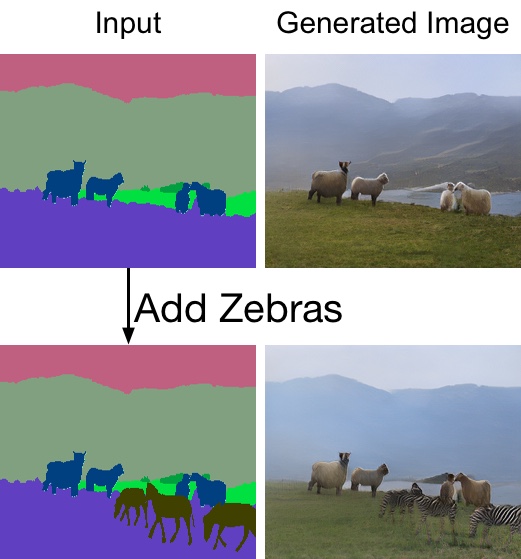}}
\subfigure[CelebAMask-HQ]{\label{fig:app_celeba}\includegraphics[width=0.355\linewidth]{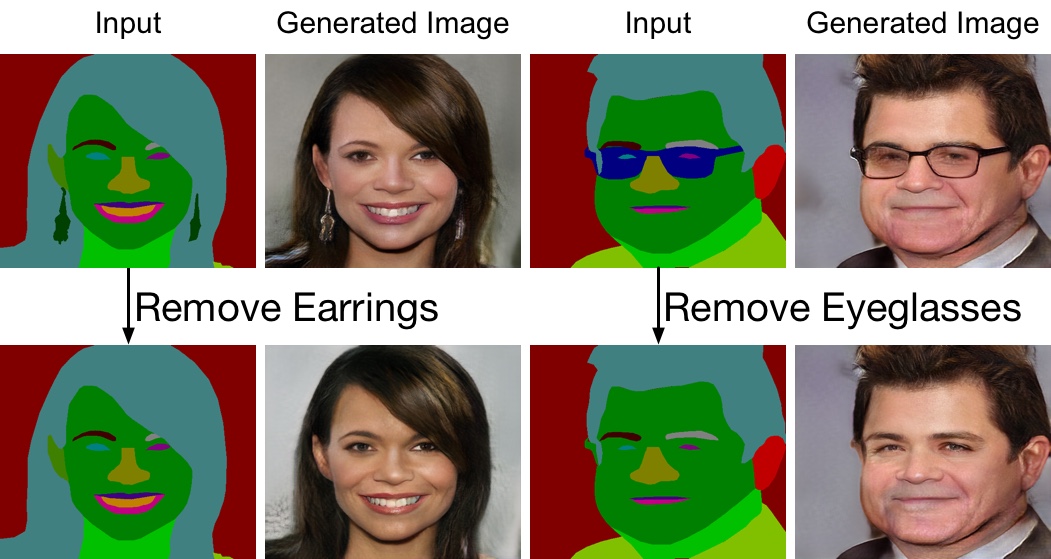}}
\subfigure[DeepFashion]{\label{fig:app_fashion}\includegraphics[width=0.635\linewidth]{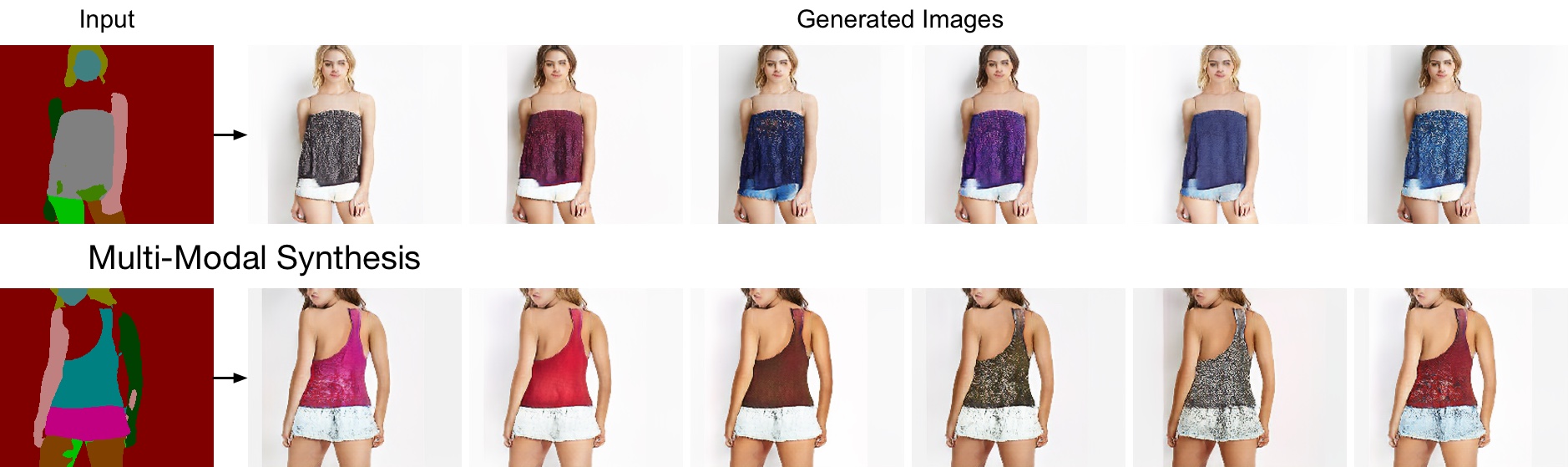}}
\caption{Exemplar applications of the proposed method on different datasets.}
\label{fig:app}
\vspace{-0.4cm}
\end{figure*}

\noindent \textbf{Semantic Manipulation.}
Our model also supports semantic manipulation.  
For instance, we can replace a building with trees (Fig.~\ref{fig:app_city}), insert a bed into a room (Fig.~\ref{fig:app_ade}), add a few zebras to some grass (Fig.~\ref{fig:app_coco}), or remove earrings and eyeglasses from a face (Fig.~\ref{fig:app_celeba}). These applications provide users more controllability during the translation process.

\noindent \textbf{Multi-Modal Synthesis.} By using a random vector as the input of the generator, our model can perform multi-modal synthesis. The results are shown in Fig.~\ref{fig:app_fashion}. We can see that our model generates different outputs from the same input layout.

